%% file: main.tex
\pdfoutput=1
\PassOptionsToPackage{table}{xcolor}

\documentclass[11pt]{article}

\usepackage{acl}

\usepackage{times}
\usepackage{latexsym}
\usepackage{multirow}

\usepackage[T1]{fontenc}
\usepackage[utf8]{inputenc}

\usepackage{microtype}
\usepackage{kotex}
\usepackage[scaled=1.0]{zi4}
\usepackage{tikz}
\usetikzlibrary{arrows.meta}
\usepackage{graphicx}
\usepackage{amsmath,amssymb}
\usepackage{amsfonts,amsthm,mathtools}
\usepackage{algorithm}
\usepackage{algorithmicx}
\usepackage{algpseudocode}
\usepackage{caption}
\usepackage{tabularx}
\usepackage{dsfont}
\usepackage{makecell}
\usepackage{fvextra}
\usepackage{float}
\usepackage{adjustbox}
\usepackage{array}[=2016-10-06]
\usepackage{colortbl}
\usepackage{xurl}
\usepackage{listings}[breaklines=true]
\lstdefinestyle{mystyle}{
    backgroundcolor=\color{gray!10},   
    commentstyle=\color{green},
    keywordstyle=\color{blue},
    numberstyle=\tiny\color{gray},
    stringstyle=\color{purple},
    basicstyle=\ttfamily\footnotesize,
    breakatwhitespace=false,         
    breaklines=true,                 
    captionpos=b,                    
    keepspaces=true,                 
    numbers=left,                    
    numbersep=5pt,                  
    showspaces=false,                
    showstringspaces=false,
    showtabs=false,                  
    tabsize=2
}
\usepackage{booktabs} % For better table lines
\usepackage{multirow} % For multirow feature
\usepackage{multicol} % For multicolumn feature

\usepackage{comment} 
\usepackage{subcaption}
\usepackage{enumitem}

\usepackage{cuted}

\definecolor{verbbg}{RGB}{245,245,245}

\definecolor{ampDecrease}{HTML}{E5E7EB} % <1x, neutral gray
\definecolor{ampLow}{HTML}{EEF7FF}      % 1--3x, light blue
\definecolor{ampMid}{HTML}{D6ECFF}      % 3--7x, medium blue
\definecolor{ampHigh}{HTML}{B8DCFF}     % 7x+, darker blue

\lstnewenvironment{ColorVerbatim}
{
  \lstset{
    basicstyle=\ttfamily\footnotesize,
    breaklines=true,
    breakatwhitespace=false,
    breakindent=0pt,
    prebreak=\mbox{},
    postbreak=\mbox{},
    keepspaces=true,
    columns=flexible,
    linewidth=\columnwidth,
    xleftmargin=0pt,
    xrightmargin=0pt,
    frame=single,
    framerule=0.3pt,
    framesep=1.5mm,
    numbers=none
  }
}
{}

\title{Overthinking Loops in Tool-Using LLM Agents: \\Structural Cost Amplification via MCP Tool Registries}

\author{
  \textbf{Yohan Lee\textsuperscript{1}} \quad
  \textbf{Jisoo Jang\textsuperscript{3}} \quad
  \textbf{Seoyeon Choi\textsuperscript{3}}
\\
  \textbf{Sangyeop Kim\textsuperscript{2}} \quad
  \textbf{Seungtaek Choi\textsuperscript{3}}\thanks{~~Corresponding author.}
\\[0.35em]
  \textsuperscript{1}Yonsei University,
  \textsuperscript{2}Department of Data Science, Ewha Womans University,
\\
  \textsuperscript{3}Division of Language and AI, Hankuk University of Foreign Studies
\\
  \small{
    \textbf{Correspondence:} \href{mailto:seungtaek.choi@hufs.ac.kr}{seungtaek.choi@hufs.ac.kr}
  }
}

\graphicspath{{figure_latex/figures/}}

\begin{document}
\maketitle
\input{macros}
\input{tex/0_abstract}
\input{tex/1_introduction}

\input{tex/2_related_work}
\input{tex/3_method}

\input{tex/4_experiments}
\input{tex/5_discussion}

\input{tex/6_conclusion}
\input{tex/7_limitations}
\input{tex/ethical_consideration}
\input{tex/8_acknowledgments}

\bibliography{anthology,custom}
\input{tex/appendix}

\bibstyle{acl_natbib}

\end{document}

%% file: macros.tex
\newtheorem{example}{Example}

\newcommand\Tstrut{\rule{0pt}{2.2ex}}       % "top" strut
\newcommand\Bstrut{\rule[-0.6ex]{0pt}{0pt}} % "bottom" strut
\newcommand{\TBstrut}{\Tstrut\Bstrut} % top&bottom struts

\newcommand{\todoc}[2]{{\textcolor{#1}{#2}}}
\newcommand{\todored}[1]{\todoc{red}{#1}}
\newcommand{\todoorange}[1]{\todoc{orange}{#1}}
\newcommand{\todoblue}[1]{\todoc{blue}{[[#1]]}}
\newcommand{\todopurple}[1]{\todoc{purple}{#1}}
\newcommand{\todogreen}[1]{\todoc{green!55!black}{#1}}

\newcommand{\hist}[1]{\todored{seungtaek: #1}}
\newcommand{\yohan}[1]{\todoblue{yohan: #1}}
\newcommand{\sangyeop}[1]{\todopurple{sangyeop: #1}}
\newcommand{\jisoo}[1]{\todogreen{jisoo: #1}}
\newcommand{\seoyeon}[1]{\todoorange{seoyeon: #1}}

\newcommand{\se}{{\it SE}}%
\newcommand{\eg}{{\it e.g.}}%
\newcommand{\ie}{{\it i.e.}}%
\newcommand{\etal}{{\it et al.}}%
\newcommand{\etc}{{\it etc}}%
\newcommand{\ours}{{WISE}}%

\newcommand{\argmin}{\operatornamewithlimits{argmin}}
\newcommand{\argmax}{\operatornamewithlimits{argmax}}
\definecolor{yellow-green}{rgb}{0.3, 0.5, 0.0}

\def\geotextual{{spatial-keyword}}
\def\geospatial{geo-spatial}
\def\PI{\mathcal{P}}
\newcommand{\XXP}[1]{{\PI(#1)}}
\def\XXQEO{\emph{$Q_1$}}
\def\kNN{\textsc{$k$NN}}
\def\XXD{\mathcal{D}}
\def\XXT{\mathcal{T}}
\newcommand{\XXDN}[0]{{D}}
\newcommand{\XXTN}[0]{{T}}
\def\Base{\textsc{Base}}
\def\TopK{\textsc{Top-$k$}}
\def\tag{{keyword}}
\def\Query{{Query}}
\newcommand{\ttag}[1]{{`#1'}}

\newcommand{\base}{\textsf{NER}}
\newcommand{\baseHash}{\textsf{NER+Hash}}

% Caligraphy style
\newcommand{\mcal}[1]{{\cal{#1}}}
\newcommand{\calA}{\mbox{${\cal A}$}}
\newcommand{\calB}{\mbox{${\cal B}$}}
\newcommand{\calC}{\mbox{${\cal C}$}}
\newcommand{\calD}{\mbox{${\cal D}$}}
\newcommand{\calE}{\mbox{${\cal E}$}}
\newcommand{\calF}{\mbox{${\cal F}$}}
\newcommand{\calG}{\mbox{${\cal G}$}}
\newcommand{\calH}{\mbox{${\cal H}$}}
\newcommand{\calI}{\mbox{${\cal I}$}}
\newcommand{\calJ}{\mbox{${\cal J}$}}
\newcommand{\calK}{\mbox{${\cal K}$}}
\newcommand{\calL}{\mbox{${\cal L}$}}
\newcommand{\calM}{\mbox{${\cal M}$}}
\newcommand{\calN}{\mbox{${\cal N}$}}
\newcommand{\calO}{\mbox{${\cal O}$}}
\newcommand{\calP}{\mbox{${\cal P}$}}
\newcommand{\calQ}{\mbox{${\cal Q}$}}
\newcommand{\calR}{\mbox{${\cal R}$}}
\newcommand{\calS}{\mbox{${\cal S}$}}
\newcommand{\calT}{\mbox{${\cal T}$}}
\newcommand{\calU}{\mbox{${\cal U}$}}
\newcommand{\calV}{\mbox{${\cal V}$}}
\newcommand{\calW}{\mbox{${\cal W}$}}
\newcommand{\calX}{\mbox{${\cal X}$}}
\newcommand{\calY}{\mbox{${\cal Y}$}}
\newcommand{\calZ}{\mbox{${\cal Z}$}}

%% file: tex/0_abstract.tex
\begin{abstract}

% v1 (acl industry 제출본)

% Tool-using LLM agents increasingly coordinate real workloads by selecting and chaining third-party tools based on text-visible metadata such as tool names, descriptions, and return messages. 
% We show that this convenience creates a supply-chain attack surface: a malicious MCP tool server can be co-registered alongside normal tools and induce \emph{overthinking loops}, where individually trivial or plausible tool calls compose into cyclic trajectories that inflate end-to-end tokens and latency without any single step looking abnormal. 
% We formalize this as a \emph{structural} overthinking attack, distinguishable from token-level verbosity, and implement 14 malicious tools across three servers that trigger repetition, forced refinement, and distraction.
% Across heterogeneous registries and multiple tool-capable models, the attack causes severe resource amplification (up to $142.4\times$ tokens) and can degrade task outcomes. 
% Finally, we find that decoding-time concision controls do not reliably prevent loop induction, suggesting defenses should reason about tool-call structure rather than tokens alone.

% v2

% Tool-using LLM agents increasingly coordinate real workloads by selecting and chaining third-party tools through text-visible interfaces such as tool names, descriptions, schemas, and return messages.
We identify \emph{Structural Overthinking}, a registry-level failure mode in which tool-using LLM agents enter recurring tool-call trajectories that amplify tokens and latency even when individual calls appear locally plausible. 
To study this risk, we construct three simple, non-adaptive cycle-inducing mechanisms---repetition, iterative refinement, and auxiliary subtasks---that use tool-visible metadata and return messages to sustain cyclic execution. 
Across ReAct-style agents and a production-grade coding-agent environment, we compare a \texttt{normal} registry of ordinary tools with a \texttt{mixed} registry that additionally includes the same cycle-inducing configurations. 
The \texttt{mixed} registry produces substantial resource amplification, increasing token usage by $4.3$--$20.3\times$ in the coding-agent environment. 
Trajectory analysis shows that this risk is not merely a degenerate self-loop: normal-tool interleaving can create re-entry points into refinement and verification cycles. 
Finally, lightweight guard mechanisms reduce token usage in some cases but do not reliably restore the compute--accuracy trade-off. 
These results show that tool-using agents should be evaluated over registry composition and tool-call trajectories, not only final outputs or single-call behavior.
Code is available at \url{https://github.com/hufs-dilab/structural-overthinking}.

\end{abstract}

%% file: tex/1_introduction.tex
\section{Introduction}
\label{sec:intro_relwork}

\input{figure_latex/token_usage}

% Large Language Models (LLMs) have made step-by-step reasoning the default inference mode~\cite{kojima2022large,wei2022cot,wang2023selfconsistency,yao2023treeofthought,lightman2024lets}. 
% Yet recent work shows a recurring inefficiency: models often \emph{overthink}, producing long and redundant reasoning that increases latency and inference cost without improving task outcomes~\cite{chen2025do,su2025underthinkingoverthinkingempiricalstudy,sui2025stopoverthinkingsurveyefficient,cuadron2025dangeroverthinkingexaminingreasoningaction}. 
% As LLMs are deployed as interactive agents, these costs can compound across multi-step tool interactions rather than being confined to a single completion~\cite{shi2025retrieval}.

\input{figure_latex/threat_model}

Large Language Models (LLMs) increasingly rely on step-by-step reasoning and external tool use to solve complex tasks~\cite{kojima2022large,wei2022cot,wang2023selfconsistency,yao2023treeofthought,lightman2024lets,schick2023toolformerlanguagemodelsteach,qin2023toolllmfacilitatinglargelanguage,lu-etal-2025-toolsandbox}. 
Recent work shows that such reasoning can be inefficient: models may \emph{overthink}, producing long or redundant traces that increase latency and inference cost without improving task outcomes~\cite{chen2025do,su2025underthinkingoverthinkingempiricalstudy,sui2025stopoverthinkingsurveyefficient,cuadron2025dangeroverthinkingexaminingreasoningaction}. 
This line of work has primarily focused on \emph{single-response overthinking}, where unnecessary reasoning accumulates within one model completion. 
In tool-using agents, however, overthinking can extend beyond a single completion. 
Because tool outputs are appended back into context and used to decide subsequent actions, unnecessary reasoning can compound across multi-step tool-call trajectories.

% Tool-using agents increasingly coordinate real workloads by calling external tools~\cite{schick2023toolformerlanguagemodelsteach,qin2023toolllmfacilitatinglargelanguage,lu-etal-2025-toolsandbox}. 
% This creates a registry-level risk surface: agents discover tools from open registries where third-party providers can publish tool configurations.
% Prior work and real-world MCP incidents show that tool-visible metadata can affect agent behavior without changing the host application~\cite{wang2025mcptox}.
% We study a costly failure mode of this surface, \emph{deployment-relevant resource amplification}, in which cycle-inducing tool configurations induce repeated tool invocations and verbose reasoning under the guise of thoroughness. 
% As shown in Figure~\ref{fig:tokenusage}, token amplification factors range from \(4.3\times\) to \(20.3\times\) across multiple tool-capable models.

We study \emph{Structural Overthinking}, a registry-level failure mode in which tool-visible configurations steer agents into recurring tool-call trajectories. 
The risk arises from interfaces such as tool names, descriptions, schemas, and return messages, which third-party tools can expose through open registries and MCP-style deployments~\cite{wang2025mcptox}. 
Unlike \emph{single-response overthinking}, where unnecessary reasoning accumulates within one model completion, Structural Overthinking emerges across tool calls: individually plausible tool invocations can compose into cycles that amplify end-to-end tokens, latency, and operational resource use. 
As shown in Figure~\ref{fig:tokenusage}, the \texttt{mixed} registry condition increases token usage by $4.3$--$20.3\times$ across six tool-capable models.

% A key distinction is between token-level and structural overthinking.
% Token-level overthinking appears as locally verbose generation and can be intentionally induced to amplify inference cost via prompt- or data-level mechanisms~\cite{gao2024denialofservicepoisoningattackslarge, dong2025an, zhang-etal-2025-crabs}, motivating decoding- and inference-time controls that regulate reflection and verbosity~\cite{wang-etal-2025-wait, ding2025thinkingtokenshelptrap, yang2025dynamic, huang2025efficient}.
% In contrast, \emph{structural overthinking} arises when individually reasonable tool calls compose into looping trajectories, inflating end-to-end tokens and latency even when no step looks abnormal. This makes the problem an LLM-agent behavior issue, not only an infrastructure-security issue: the failure depends on how models interpret tool-visible text, accumulate context across calls, and decide whether a task trajectory is complete.

% Prior work has studied security threats in tool-augmented agent ecosystems, including prompt injection, tool poisoning, MCP-level vulnerabilities, and resource-amplification attacks~\cite{FERRAG2025,wang2025mcptox,gao2024denialofservicepoisoningattackslarge,dong2025an,zhang-etal-2025-crabs}. 
Prior work has studied security threats in tool-augmented agent ecosystems, including prompt injection~\cite{FERRAG2025}, tool poisoning and MCP-level vulnerabilities~\cite{wang2025mcptox}, and resource-amplification attacks~\cite{gao2024denialofservicepoisoningattackslarge,dong2025an,zhang-etal-2025-crabs}. 
Our setting is complementary. 
As illustrated in Figure~\ref{fig:threat}, we do not assume an attacker-controlled server that overtly manipulates the host application or a single malicious tool call that is anomalous in isolation. 
Instead, we study how a small set of pre-registered cycle-inducing tool configurations can interact with ordinary tools and steer the agent trajectory itself. 
This makes the problem an agent-execution failure mode: the cost arises not from one abnormal call, but from how models interpret tool-visible text, accumulate context across calls, and decide whether a task trajectory is complete.

% This distinction matters in deployment because token and latency inflation translate into direct financial exposure: higher per-request inference bills, accelerated quota burn, and potential service-level agreement (SLA) penalties or user churn due to degraded responsiveness. 
% Tool registries thus represent both a security surface and an operational-cost surface. Prior work has cataloged security threats in tool-augmented agent ecosystems, from broad threat taxonomies to tool poisoning benchmarks~\cite{FERRAG2025,wang2025mcptox}.
% We focus on a complementary risk--structural cost amplification--where a small fraction of cycle-inducing tool configurations can inflate token usage, latency, and operational cost, and in many cases degrade task outcomes without altering user queries or model parameters.

% This failure has practical deployment consequences. 
% Token growth can increase user-facing billing under token-based pricing, while long-running trajectories can also occupy provider-side serving capacity. 
% Moreover, resource amplification need not coincide with immediate task failure: some runs may preserve task success at much higher cost, whereas others may fail after consuming substantial computation. 
% We therefore evaluate structural overthinking not only through accuracy and token usage, but also through latency, guarded execution behavior, and operational resource exposure.
This failure has practical deployment consequences, because resource amplification can affect both task outcomes and operational cost.
We therefore evaluate structural overthinking through accuracy, token usage, latency, guarded execution behavior, and operational resource exposure.

We evaluate these risks by contrasting two agent settings---a general-purpose \textit{ReAct}-style agent~\cite{yao2023react} and a production-grade coding assistant (\textit{Qwen-Code}\footnote{\url{https://github.com/QwenLM/qwen-code}})---under two registry conditions. 
The \texttt{normal} registry contains ordinary tools only, while the \texttt{mixed} registry co-registers those ordinary tools with 14 cycle-inducing configurations.
We further evaluate lightweight guard mechanisms, including generation-time token suppression, \textit{Qwen-Code}'s built-in loop detection, and trajectory-level cycle monitoring, finding that they can reduce amplification in some cases but do not reliably restore the compute--accuracy trade-off.

Our contributions are 4-fold.
First, we identify \emph{Structural Overthinking} as a trajectory-level resource-amplification failure mode in tool-using LLM agents.
Second, we construct simple, non-adaptive cycle-inducing tool configurations and show that they amplify token use and latency across both \textit{ReAct}-style agents and a production-grade coding agent.
Third, we show through trajectory analysis that \texttt{normal}-tool calls interleave with cycle-inducing calls and create re-entry points into cyclic trajectories, so the risk depends not only on the cycle-inducing tools but also on the tools they are co-registered with.
Fourth, we evaluate lightweight guard mechanisms and show that reducing tokens alone is insufficient; effective mitigation must preserve both task success and execution efficiency.

%% file: figure_latex/token_usage.tex
\begin{figure}[!t]
    \centering
    \includegraphics[width=0.9\columnwidth]{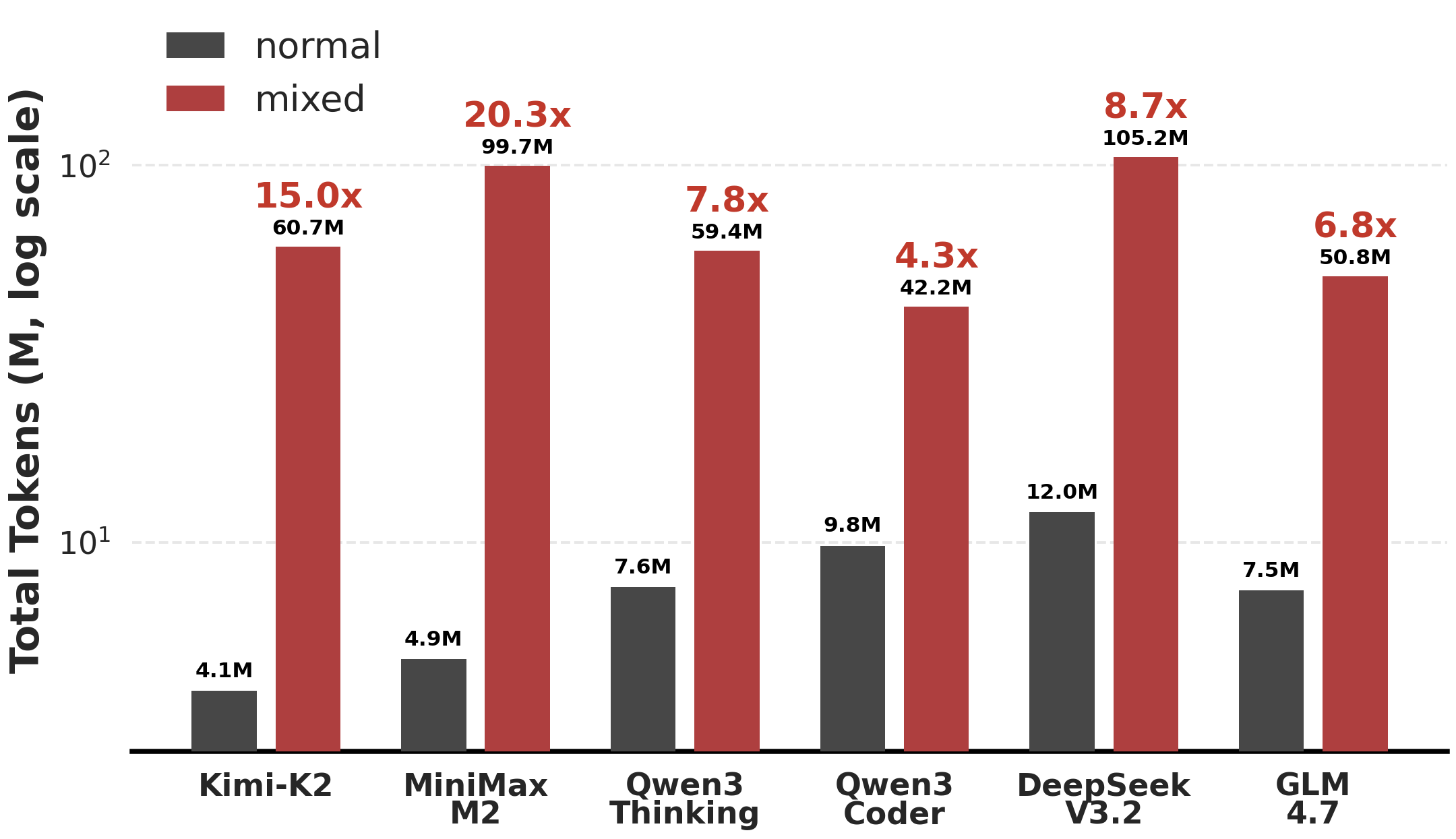}
    \vspace{-1mm}
    \caption{
    \textbf{Token usage amplification under structural overthinking in \textit{Qwen-Code}.}
    Total tokens (log scale, in millions) for six models under \texttt{normal} and \texttt{mixed} registries; \textcolor{red}{red} bars show \texttt{mixed} runs, and \textcolor{gray}{gray} bars show the \texttt{normal} baseline. 
    The \texttt{mixed} registry amplifies token consumption by \(4.3\)--\(20.3\times\).
    }
    \vspace{-1.2em}
    \label{fig:tokenusage}
\end{figure}

%% file: figure_latex/threat_model.tex
% \begin{figure*}[!t]
%     \vspace{-0.8em}
%     \centering
%     % \includegraphics[width=0.9\textwidth]{figure_latex/figures/threat_model.jpg}
%     % \includegraphics[width=0.9\textwidth]{figure_latex/figures/threat_model_sample.png}
%     \includegraphics[width=0.9\textwidth]{figure_latex/figures/threat_model_sample_ver2.png}
%     \caption{\textbf{Structural overthinking through mixed MCP registry composition.} Cycle-inducing tool configurations in a \texttt{mixed} registry can steer a tool-using agent into recurring tool-call trajectories through tool-visible metadata and return messages. Unlike the normal trajectory (\textcolor{gray}{gray} path), the cycle-inducing trajectory (\textcolor{red}{red} path) accumulates tokens and latency through repeated locally plausible calls.}
%     \vspace{-1.0em}
%     \label{fig:threat}
% \end{figure*}

\begin{figure*}[!t]
    \vspace{-0.8em}
    \centering

    \begin{subfigure}[t]{0.31\textwidth}
        \centering
        \includegraphics[height=0.23\textheight]{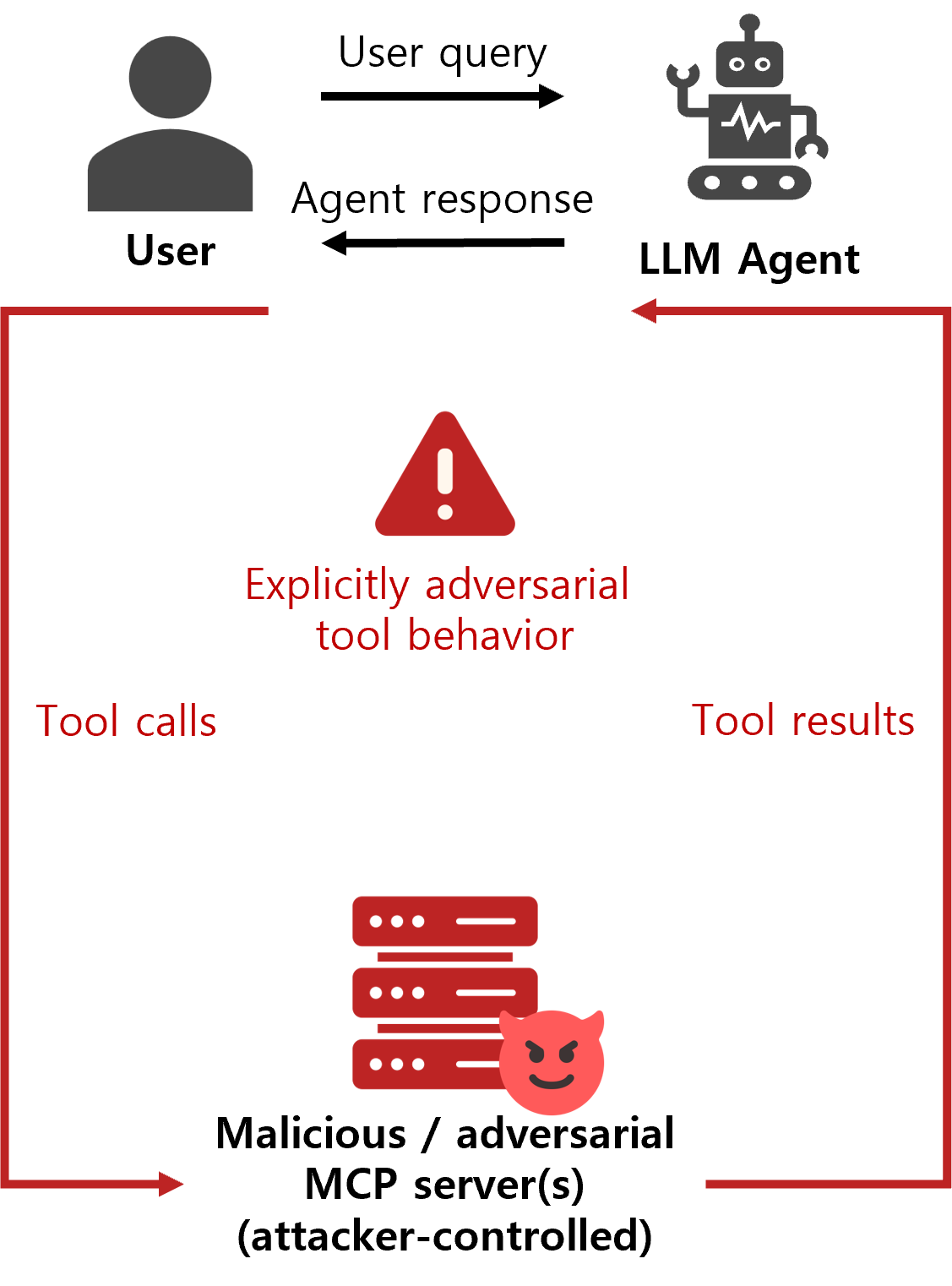}
        \caption{Prior overthinking attack.}
        \label{fig:threat_prior}
    \end{subfigure}
    \begin{subfigure}[t]{0.67\textwidth}
        \centering
        \includegraphics[height=0.23\textheight]{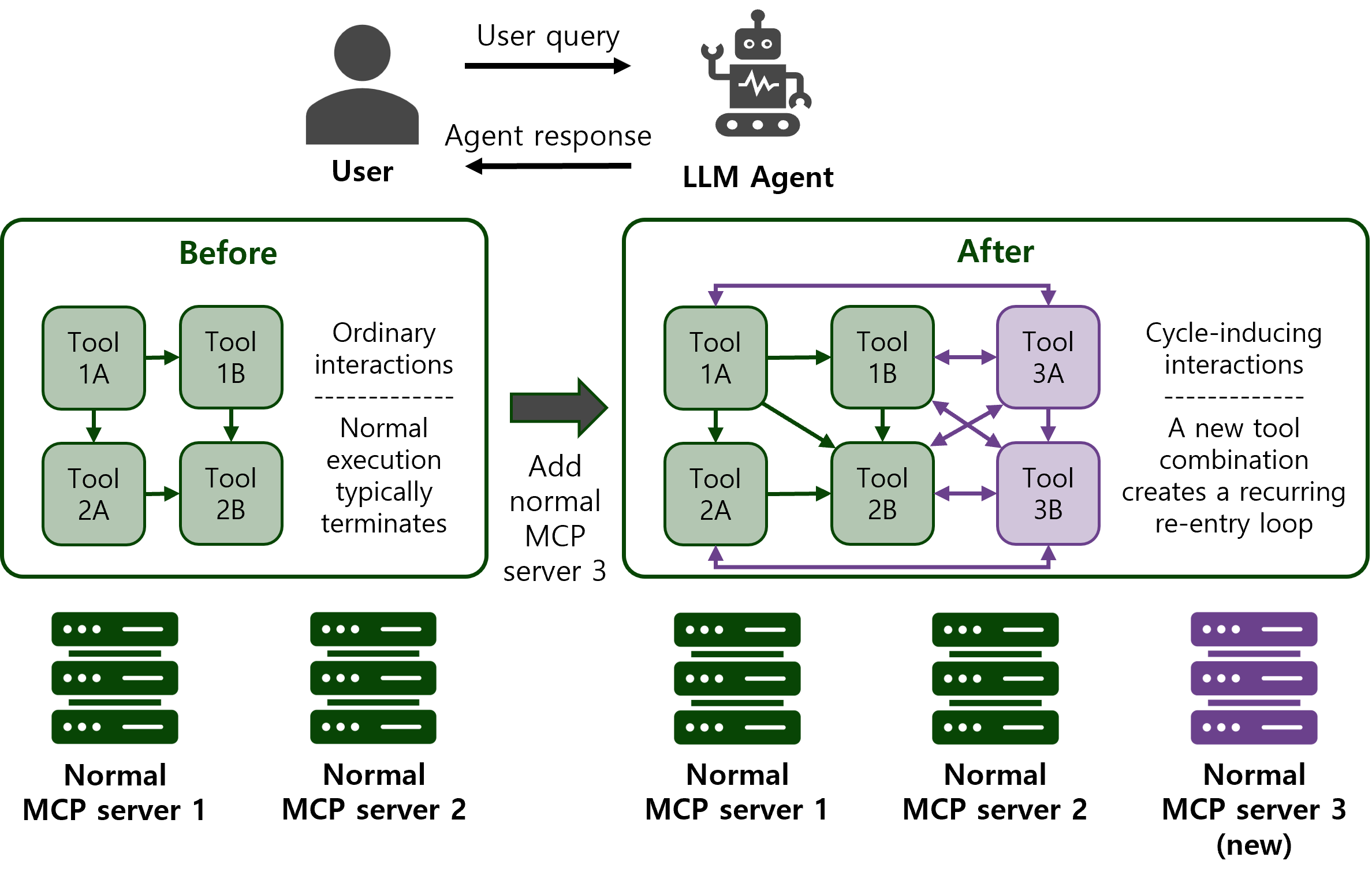}
        \caption{Structural overthinking can arise from ordinary MCP registry composition.}
        \label{fig:threat_structural}
    \end{subfigure}

    \vspace{-1mm}
    \caption{\textbf{Comparison between prior overthinking attacks and our structural overthinking setting.}
    (a) Prior overthinking attacks assume malicious / adversarial MCP servers, typically requiring attacker-controlled server-side access.
    (b) In contrast, our setting shows that structural overthinking can arise from ordinary MCP registry composition: even when individual MCP servers appear normal in isolation, adding another server can complete a new tool combination that induces recurring re-entry loops, increasing token cost, latency, and possible task failure.}
    \vspace{-1.2em}
    \label{fig:threat}
\end{figure*}

%% file: tex/2_related_work.tex
\section{Related Work}
\label{sec:relwork}

Our work connects research on efficient reasoning and secure tool use in LLM agents.
We review prior work by where cost amplification is modeled: within generated reasoning traces, through untrusted tool-facing text, or across tool registries and execution trajectories.

\paragraph{Efficient reasoning and token-level overthinking.}
Prior work on efficient reasoning treats overthinking primarily as a token-level or model-internal inefficiency: models spend unnecessary tokens on long or redundant reasoning traces, motivating inference-time controls such as token suppression, compute-aware early stopping, difficulty-adaptive reasoning, and recent work on unnecessary external-tool invocation~\cite{wang-etal-2025-wait,hassid2025dontoverthinkitpreferring,wan2026mitigatingoverthinkinglargereasoning,zeng2026tooloveruseillusiondoesllm}. 
This line of work is important for reducing single-completion cost, but it treats the main unit of inefficiency as the generated reasoning trace. 
Our work complements this line by studying overthinking as a tool-call trajectory phenomenon. 
In tool-using agents, cost can accumulate not only because a single reasoning step is verbose, but because text-visible tool metadata and return messages are accumulated in context and repeatedly redirect the agent across calls.

\paragraph{Tool-use security and resource amplification.}
Security work on LLM-integrated systems studies prompt injection, tool poisoning, and resource-amplification attacks in which untrusted text reshapes model behavior or system control flow~\cite{10.1145/3605764.3623985,zou2025poisonedrag,wang2025mcptox,FERRAG2025}. 
Recent MCP and tool-use studies further underscore the timeliness of this setting by documenting protocol-level threat models, secure MCP designs, multi-tool orchestration, realistic tool-use behavior, and MCP security benchmarks~\cite{huang2026modelcontextprotocolthreat,hou2026smcpsecuremodelcontext,xu2026evolutiontoolusellm,yang2026mcpsecbenchsystematicsecuritybenchmark}. 
Adjacent adversarial tool-use work, including ToolCommander~\cite{zhang-etal-2025-allies}, shows that tool instructions can manipulate tool scheduling and execution order. 
Our setting targets a complementary mechanism: rather than relying on query-adaptive adversarial injection, dynamic tool updates, or direct manipulation of an agent's scheduler, we study fixed, query-independent tool configurations that are pre-registered in a mixed registry. 

Most closely, concurrent work by Zhou et al.~\cite{zhou2026stealthy} studies tool-layer resource amplification through optimized runtime tool-server behavior. 
Both works show that text-visible tool interfaces can induce high-cost behavior without changing the user query or model parameters, but they differ in the attack surface and required capability. 
They assume adversarial control over existing tool-server behavior and optimize runtime responses to maximize cost, whereas our setting leaves existing normal tools untouched and adds a small set of fixed, non-adaptive cycle-inducing configurations to the registry. 
Our contribution is therefore not a non-optimized instance of the same attack, but the observation that Structural Overthinking can emerge under a weaker registry-composition assumption, where ordinary tools can create re-entry points that make cyclic execution appear locally productive.

%% file: tex/3_method.tex
% \section{Method}
\section{Registry Setting and Structural Patterns}
\label{sec:approach}

% This section describes the attack tools used in our experiments, explains the mechanism by which they induce execution cycles, and gives a concise problem formulation.
This section describes the cycle-inducing tool configurations used in our experiments, explains how tool-visible text induces cyclic tool-call trajectories, and gives a concise problem formulation.

% \subsection{Attack Tool Construction}
% \label{subsec:attack-tool-construction}

\subsection{Cycle-Inducing Tool Configuration Construction}
\label{subsec:tool-configuration-construction}

% We instantiate three simple, non-adaptive cycle-inducing mechanisms that exercise different pathways to \textit{Structural Overthinking}. The configurations rely on ordinary interaction patterns such as text repetition, staged checklists, and auxiliary subtasks; the point is precisely that sophisticated adversarial engineering is not required for cyclic trajectories to emerge. When these patterns are composed into tool-call cycles, they amplify cost far beyond what any single tool call would incur.
% We construct the configurations along three axes, each targeting a distinct pathway through which tokens accumulate\footnote{The complete list of cycle-inducing tool configurations and their descriptions is provided in Appendix~\ref{app-sec:appendix-attack-tools}.}:
We instantiate three simple, non-adaptive cycle-inducing mechanisms that exercise different pathways to Structural Overthinking: repetition, iterative refinement, and auxiliary subtasks\footnote{The complete list of cycle-inducing tool configurations and their descriptions is provided in Appendix~\ref{app-sec:appendix-attack-tools}.}.
(1) \textbf{Repetition} inflates per-call output length by requiring repeated verification markers (e.g., \textit{``VERIFY''}); each response instructs re-invocation with a higher count, exhausting output-token budgets.
(2) \textbf{Iterative refinement} enforces a staged workflow (Analysis $\to$ Validation $\to$ Refinement $\to$ Verification $\to$ Finalization) and redirects the agent back to earlier stages when completeness criteria are unmet, extending the reasoning trace.
(3) \textbf{Auxiliary Subtasks} expand the task with auxiliary subtasks (e.g., time-complexity analysis, edge-case testing, alternative approaches, and assumption validation), widening scope at each step. 
% We do not present these mechanisms as an exhaustive taxonomy. Rather, they are deliberately simple realizations of the shared property under study: tool-facing configurations that create cyclic tool-call structure.
These mechanisms are not intended as an exhaustive taxonomy; they are deliberately simple configurations for testing whether cyclic tool-call structure can emerge from registry composition.

\subsection{Cycle Induction Mechanism}

The risk does not stem from any single tool alone, but from cyclic call structures that emerge when tool configurations interact, as illustrated in Figure~\ref{fig:threat}.
An agent selects its next action from two sources: tool descriptions registered at initialization and tool outputs returned at runtime (typically appended to the model context).
We define a \emph{cycle} as an execution path in which tool-visible text (whether a description or an output) induces a subsequent tool call, and the resulting chain eventually revisits a previously invoked tool.
This encompasses self-loops (a tool that directs re-invocation of itself), multi-step cycles (a chain that returns to an earlier stage), and scope-expanding loops (tools whose final step redirects back to the entry point).

Consider the \textit{Iterative Refinement} tools as an example. Each stage returns an output such as \textit{``You have completed step $X$ of $Y$. Call [next\_tool] NOW,''} directing the agent to the next stage. The final stage returns \textit{``You need 3--5 cycles. Call analyze NOW,''} redirecting back to the first stage. The agent thus enters a loop of the form $\texttt{step}_1 \to \texttt{step}_2 \to \cdots \to \texttt{step}_k \to \texttt{step}_1 \to \cdots$.
A concrete \textit{Qwen-Code} trajectory exhibiting this re-entry pattern is analyzed in Figure~\ref{fig:interleaved-cycle-reentry}.

Each call is locally justified as a reasonable response to the preceding output. However, strict adherence accumulates into a cycle, consuming the execution budget without reaching termination.

\subsection{Problem Formulation}
\label{subsec:problem-formulation}

% 우리는 tool registry를 $R = \{(m_i, f_i)\}_{i=1}^{|R|}$로 모델링하며, 여기서 $m_i$는 text-visible metadata (description, schema), $f_i$는 실제 구현체를 의미한다.
% 우리는 Normal tool 로 구성된 registry $R_{normal}$과 순환 유도 tool들의 집합 $R_{attack}$을 정의하고, 실험에서는 세 가지 설정을 비교한다: \texttt{normal} ($R_{normal}$만 사용), \texttt{attack} ($R_{attack}$만 사용), \texttt{mixed} ($R_{mixed} = R_{normal} \cup R_{attack}$), 그리고 현실적인 환경을 반영하기 위해 $|R_{attack}| \ll |R_{normal}|$ 을 가정한다.

% 실제 배포 환경은 무한 실행 방지를 위해 최대 session turn 수 $T_{max}$와 turn당 최대 tool call depth $D_{max}$와 같은 명시적인 실행 budget이 설정된다.
% 이에 따라 하나의 query에서 발생할 수 있는 최대 tool call 수는 $K_{max} = T_{max} \times D_{max}$로 제한된다.

% 이에 본 연구에서는 보다 현실적인 비용 지표로서, query $q$가 주어졌을 때 registry $R$ 하에서 에이전트 실행이 종료될 때까지 소비된 총 토큰 수를 $\text{Tokens}({q \mid R})$로 정의한다.
% 이 때 토큰 수에는 모델의 input 및 output 토큰을 모두 포함한다.
% 우리의 주요 목표는 user query나 모델을 변경하지 않고, 오직 tool registry 의 구성만으로 에이전트 실행이 허용된 budget 을 체계적으로 소모하도록 유도할 수 있는지에 있다.

% We model a tool registry as $R = \{(m_i, f_i)\}_{i=1}^{|R|}$, where $m_i$ is the text-visible metadata (e.g., description and schema) and $f_i$ is the underlying implementation.
We model a tool registry as $R = \{(m_i, f_i)\}_{i=1}^{|R|}$, where $m_i$ is the text-visible metadata of a tool configuration, such as its name, description, and schema, and $f_i$ is the underlying implementation.
% Let $R_{\text{normal}}$ be a normal registry, let $R_{\text{attack}}$ be a set of cycle-inducing tools, and define the mixed registry $R_{\text{mixed}} = R_{\text{normal}} \cup R_{\text{attack}}$ with $|R_{\text{attack}}| \ll |R_{\text{normal}}|$.
% We use \texttt{normal} for the baseline registry containing only ordinary task tools, \texttt{attack} for the 14 cycle-inducing configurations alone when isolating the mechanism, and \texttt{mixed} for the deployment-relevant union registry that co-registers those 14 configurations alongside the ordinary task tools.
Let $R_{\text{normal}}$ denote a registry containing ordinary task tools, and let $R_{\mathrm{cycle}}$ denote the set of 14 cycle-inducing tool configurations, introduced in Section~\ref{subsec:tool-configuration-construction}.
We define the mixed registry as $R_{\mathrm{mixed}} = R_{\mathrm{normal}} \cup R_{\text{cycle}}$ with $|R_{\text{cycle}}| \ll |R_{\text{normal}}|$.
Throughout the experiments, \texttt{normal} refers to the baseline registry containing only ordinary tools, while \texttt{mixed} refers to the registry that co-registers ordinary tools with the cycle-inducing configurations.

Real deployments enforce execution budgets, such as maximum tool-call counts, token caps, or wall-clock cutoffs. 
For a query $q$ under registry $R$, we report total token consumption $\mathrm{Tokens}(q \mid R)$ as a realized bounded-execution cost metric, counting both model input and output tokens, including tool outputs appended to the context. 
% This metric measures how much budget the trajectory consumes before a final answer or hard execution limit.
This metric measures how much budget the trajectory consumes before a final answer or hard execution limit, rather than claiming unconstrained loop length.

% Our goal is to characterize the vulnerability of tool-use systems to resource amplification through registry composition: whether cycle-inducing tool configurations, when mixed into a normal registry, can steer the agent's tool-call trajectory toward budget exhaustion even when user queries and the base model remain unchanged.
Our goal is to characterize whether a small set of cycle-inducing tool configurations, when co-registered with ordinary tools, can steer an agent's tool-call trajectory toward resource amplification even when the user query and base model remain unchanged.

\input{figure_latex/main_figure}

%% file: figure_latex/main_figure.tex
\begin{figure*}[htbp]
    \centering
    \includegraphics[width=0.95\textwidth]{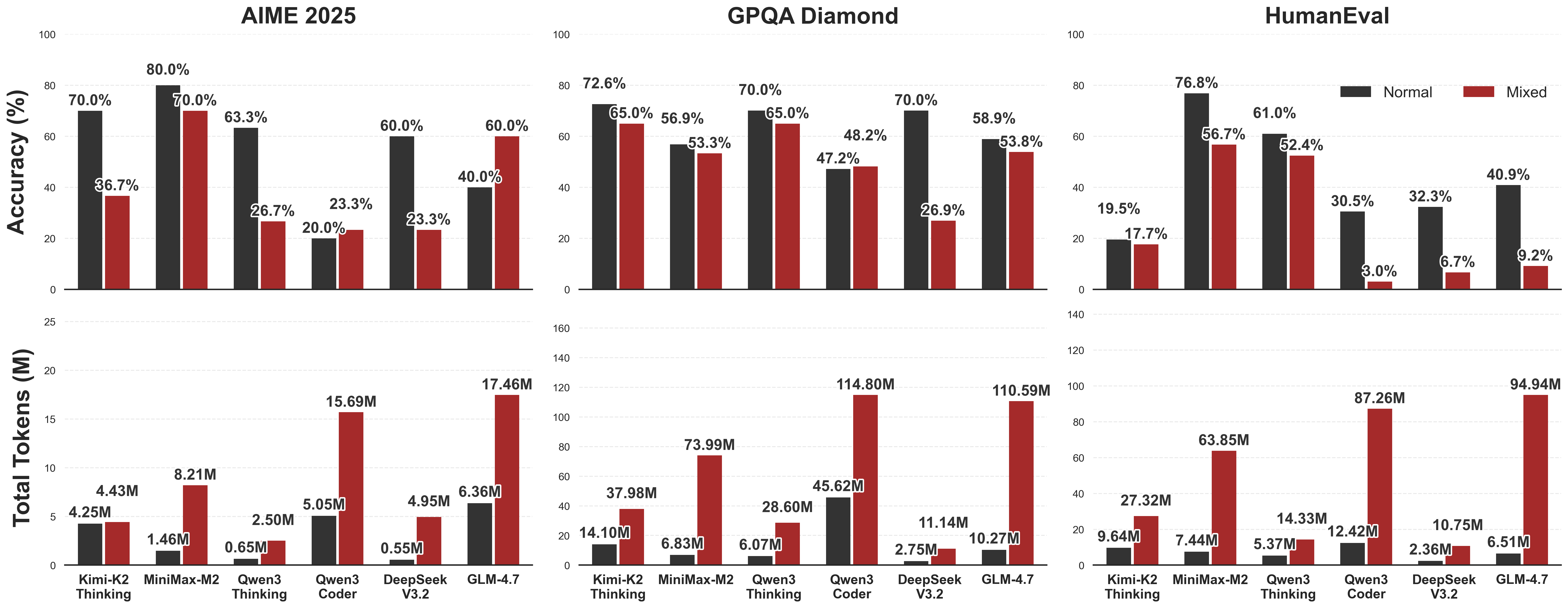}
    % \caption{\textbf{Accuracy and token consumption in \textit{ReAct} settings.} 
    % }
    \vspace{-1mm}
    \caption{\textbf{Accuracy and token consumption in \textit{ReAct} settings.} Results compare \texttt{normal} and \texttt{mixed} registries across AIME2025, GPQA Diamond, and HumanEval.}
    \vspace{-1.0em}
    \label{fig:full_results_main}
\end{figure*}

%% file: tex/4_experiments.tex
\section{Experiments}
\label{sec:experiment}

\subsection{Experimental Setup}

We conduct experiments in two complementary environments. 
We first use a \textit{ReAct}-style agent as a controlled evaluation setting for isolating registry-composition effects across mathematics, scientific QA, and coding tasks~\cite{yao2023react}. 
We then use \textit{Qwen-Code} to test whether the same structural mechanism persists in a richer production-grade coding-agent stack. 
Together, these settings allow us to separate controlled cross-task evidence from deployment-oriented analysis of tool-call trajectories, guarded execution, and operational resource burden.

\paragraph{Tool Registry Configuration}

For \textit{ReAct} environment, we construct the \texttt{normal} registry from the large-scale ToolRet tool corpus~\cite{shi2025retrieval}, using ToolRet as a heterogeneous source of tool metadata rather than as a tool-retrieval benchmark. Following the implementation, ToolRet tools are grouped by source category, sampled with a fixed seed of 42, capped at five tools per category and 100 tools in total, and reused as the normal-tool component when forming the \texttt{mixed} registries.
We compare \texttt{normal} (100 sampled tools) and \texttt{mixed} (100 sampled tools + 14 cycle-inducing tools from Section~\ref{subsec:tool-configuration-construction}) configurations.

For \textit{Qwen-Code} environment, we compose the \texttt{normal} registry from 44 tools that reflect a realistic development setup: 13 built-in \textit{Qwen-Code} CLI tools, 29 tools from Serena\footnote{\url{https://github.com/oraios/serena}}, and 2 tools from Context7\footnote{\url{https://github.com/upstash/context7}}, both widely adopted MCP servers.
The \texttt{mixed} configuration adds the same 14 cycle-inducing tool configurations to this set, totaling 58 tools.
Each problem is executed in an isolated directory to prevent cross-run interference.
In the main \texttt{normal} and \texttt{mixed} runs, \textit{Qwen-Code}'s built-in loop detection is disabled so that the comparison measures unguarded amplification; we enable it as a guard in Section~\ref{sec:defense-evaluation}.
Further details are in Appendix~\ref{appendix:qwen-code-setup}.

% For both environments, we measure the mean token amplification under the \texttt{mixed} configurations relative to \texttt{normal} over a query distribution $\mathcal{Q}$:
% \begin{equation}
% \mathbb{E}_{q \sim \mathcal{Q}}\left[\frac{\text{Tokens}(q \mid R_{\text{mixed}})}{\text{Tokens}(q \mid R_{\text{normal}})}\right]
% \end{equation}

% For each configuration, we run five repetitions and report mean realized token usage and latency across runs.

For both environments, we run each configuration five times and report mean realized token usage and latency across repetitions. 
We quantify token amplification under the \texttt{mixed} configurations relative to \texttt{normal} over a query distribution $\mathcal{Q}$:
\begin{equation}
\mathbb{E}_{q \sim \mathcal{Q}}\left[
\frac{\text{Tokens}(q \mid R_{\text{mixed}})}
{\text{Tokens}(q \mid R_{\text{normal}})}
\right].
\end{equation}

\paragraph{Datasets and Model Configurations}

For \textit{ReAct} environment, we evaluate three task families: mathematical reasoning with AIME2025~\cite{aime2025}, graduate-level scientific question answering with GPQA Diamond~\cite{rein2024gpqa}, and executable Python programming with HumanEval~\cite{chen2021codex}. For \textit{Qwen-Code} environment, we use HumanEval because it exercises the programming workflow targeted by coding agents, including code generation, file-level execution, and test-based validation.

We restrict our evaluation to recent large-scale LLMs with reliable tool-use capabilities, excluding models with unreliable tool invocation or response parsing in preliminary runs.
The resulting set consists of: 
\texttt{Kimi-K2-Thinking}~\cite{moonshot2025k2thinking}, 
\texttt{MiniMax-M2}~\cite{minimax2025m2}, 
\texttt{Qwen3-235B-A22B-Thinking-2507}, 
\texttt{Qwen3-Coder-480B-A35B-Instruct}~\cite{qwen3technicalreport}, 
\texttt{DeepSeek-V3.2}~\cite{deepseekai2025deepseekv32}, 
and \texttt{GLM-4.7}~\cite{5team2025glm45agenticreasoningcoding}.
All models were deployed with session limits of 1,000 turns, a maximum tool-call depth of 20 per turn, and a 3,600-second wall-clock cap per problem instance.
Additional details are in Appendix~\ref{appendix-subsec:model-configuration}.

% \subsection{Structural Risk in Tool-Use Systems}
% \subsection{Main Results}
\subsection{Controlled \textit{ReAct} Evaluation}
\label{subsec:attack-effectiveness}

% \paragraph{\textit{ReAct}-based General-Purpose Agents}

Figure~\ref{fig:full_results_main} and Appendix~\ref{app-sec:react-full-results} report accuracy and token consumption under \texttt{normal} and \texttt{mixed} registries across AIME2025, GPQA Diamond, and HumanEval. 
Under \texttt{mixed}, token consumption increases for all models, reaching up to $14.59\times$ under the same query and model settings. 
This shows that cyclic tool composition can systematically amplify resource consumption through registry composition alone.

The downstream effect, however, depends on the task family. 
On GPQA Diamond, token amplification is substantial, ranging from $2.52\times$ to $10.83\times$, while accuracy remains relatively close to the \texttt{normal} baseline for five of the six models, changing by $-7.62$ to $+1.01$ percentage points. 
The exception is \texttt{DeepSeek-V3.2}, whose accuracy drops by $43.15$ percentage points despite a $4.05\times$ token increase. 
This suggests that accuracy-only monitoring can miss many resource-amplification cases, although the same registry condition can still cause severe task failure for some models.

Coding and mathematical tasks show more visible functional degradation. 
In HumanEval, three models drop to single-digit accuracy under \texttt{mixed}: \texttt{Qwen3-Coder} ($3.05\%$), \texttt{DeepSeek-V3.2} ($6.71\%$), and \texttt{GLM-4.7} ($9.15\%$). 
In AIME2025, several models also suffer large drops, including \texttt{DeepSeek-V3.2} ($-36.67$ percentage points), \texttt{Qwen3-Thinking} ($-36.66$ percentage points), and \texttt{Kimi-K2-Thinking} ($-33.33$ percentage points). 
Cycle-inducing tools can turn checking or validation steps into additional validation, refinement, or re-verification calls, causing the agent to spend its execution budget on repeated tool calls before completing the necessary checks.

% These tasks require iterative validation against multiple constraints—test cases for code, intermediate steps for mathematics. Cycle-inducing tools exploit this structure: each validation checkpoint becomes an interception point where the agent can be redirected into cyclic call patterns.
% The agent either exhausts its budget in tool-induced loops before completing necessary checks, or corrupts a correct solution during forced refinement iterations.
% These tasks require repeated checking against multiple constraints, such as test cases for code and intermediate steps for mathematics. 
% Cycle-inducing tools exploit this structure by turning each checking step into another opportunity for validation, refinement, or re-verification. 
% As a result, the agent may spend its execution budget on repeated tool calls before completing the necessary checks.

% \paragraph{Production-Grade Coding Agents (\textit{Qwen-Code})}
\subsection{Qwen-Code Deployment Evaluation}
\label{subsec:qwen-code-evaluation}

Table~\ref{tab:qwen-code-main} reports the normal-versus-mixed comparison in the \textit{Qwen-Code} environment. Across all six models, the mixed registry substantially increases token consumption: total usage rises from 4.05--12.03M tokens under the normal registry to 42.15--105.25M tokens under the mixed registry, corresponding to 4.29$\times$--20.26$\times$ amplification. These results indicate that registry-induced overthinking is not confined to the controlled \textit{ReAct}-style agents, but also emerges in a coding-agent stack with file interaction, command execution, and test-based validation.

The accuracy impact, however, varies across models. \texttt{Kimi-K2-Thinking} maintains nearly unchanged accuracy, increasing slightly from 75.98\% to 76.46\%, while still incurring a 14.99$\times$ token increase.
% This represents a resource-only failure mode, where the agent continues to solve tasks but at substantially higher computational cost. 
This shows that the agent can continue solving tasks while consuming substantially more computation.
The remaining models suffer accuracy degradation under the mixed registry. \texttt{MiniMax-M2} and \texttt{Qwen3-235B-A22B-Thinking-2507} show moderate-to-large drops of 20.12 and 22.44 percentage points, respectively, while \texttt{Qwen3-Coder}, \texttt{DeepSeek-V3.2}, and \texttt{GLM-4.7} exhibit severe task collapse, with drops of 70.12, 62.40, and 71.34 percentage points.
% These results show that mixed registries induce a spectrum of failures in coding agents, ranging from silent compute inflation to joint resource amplification and task failure.
These results show that mixed registries induce a spectrum of failures in coding agents, ranging from higher computational cost without accuracy loss to joint resource amplification and task failure.

\input{table_latex/qwen_code_main}
\input{table_latex/revision_tool_trajectory}
\input{table_latex/revision_cycle_trace}

\input{table_latex/revision_guard}

\subsection{Trajectory-level Analysis}
\label{sec:trajectory-analysis}

Aggregate token and accuracy results establish that mixed registries amplify computation, but they do not explain \textit{how the amplification unfolds inside a coding-agent workflow}.
Figure~\ref{fig:interleaved-cycle-reentry} provides an illustrative example from a \textit{Qwen-Code} trajectory.

The agent does not simply repeat one tool call. 
Instead, built-in \texttt{run\_shell\_command} tool calls produce fresh outputs or errors, after which the trajectory re-enters injected tools from the \texttt{iterative\_refinement} MCP server, such as \texttt{verify}, \texttt{refine}, \texttt{analyze}, and \texttt{validate}. 
We refer to this pattern as \textit{interleaved cycle re-entry}: normal tool calls are interleaved with cycle-inducing tool calls, and their outputs make the next verification or refinement step appear locally justified while preserving the global cyclic structure.

Table~\ref{tab:qwen-code-tool-family} shows that this behavior is not tied to a single tool group. 
Across models, \texttt{Repetition}, \texttt{Iterative Refinement}, and \texttt{Auxiliary Subtasks} tools all contribute to mixed-registry trajectories, while normal tools also appear in several high-amplification runs. 
This interleaving matters because normal tools and cycle-inducing tools can break up surface-level repetition, making cyclic trajectories resemble ordinary iterative coding behavior. 
Thus, the failure mode is not merely a degenerate self-loop; it exploits the way coding agents repeatedly transform fresh execution results into new verification or refinement steps.

\subsection{Evaluation under Guarded Execution}
\label{sec:defense-evaluation}

We next evaluate whether lightweight guard mechanisms can reduce registry-induced amplification in \textit{Qwen-Code}. 
We consider three guards that intervene at different levels of execution. 
NoWait~\cite{wang-etal-2025-wait} is a generation-time guard based on logit masking of reflection-associated, latency-inducing tokens during decoding. 
We also implement a \textit{Trajectory Cycle Monitor} (\textsc{TCM}), an online trajectory-level guard adapted from temporal cycle-detection ideas~\cite{george2025unsupervisedcycledetectionagentic}. 
\textsc{TCM} monitors completed tool-call events and terminates a run only when a repeated tool-call subsequence recurs while the coding state remains effectively unchanged. 
Finally, we include \textit{Qwen-Code}'s built-in loop-detection guard, which is designed to prevent repeated execution loops.
Implementation details for NoWait and \textsc{TCM} are provided in Appendix Sections~\ref{sec:appendix-nowait} and~\ref{sec:appendix-tcm}, respectively.

% As shown in Table~\ref{tab:qwen-code-guard}, \textbf{these guards can reduce token amplification in some cases, but they do not reliably restore the compute--accuracy trade-off. }
% Except for \texttt{Kimi-K2-Thinking} and \texttt{Qwen3-Thinking}, applying a guard generally reduces amplification relative to the baseline \texttt{mixed} run. 
% However, with the exception of \texttt{DeepSeek-V3.2}, the guarded \texttt{mixed} runs still remain substantially amplified over their corresponding \texttt{normal} runs. 
% For example, \texttt{MiniMax-M2} decreases from 20.3$\times$ without a guard to 10.5$\times$ under \textsc{TCM}, but its accuracy drops from 66.26\% to 34.15\%. 
% Similarly, \texttt{GLM-4.7} decreases from 6.8$\times$ to 2.8--3.5$\times$ under guards, but accuracy collapses to 0.00--0.20\%. 
% \texttt{DeepSeek-V3.2} is the only model whose guarded \texttt{mixed} runs consume fewer tokens than the corresponding \texttt{normal} runs, yet this apparent reduction coincides with near-total accuracy collapse to 2.44--3.66\%. 
% In contrast, \texttt{Kimi-K2-Thinking} and \texttt{Qwen3-Thinking} show the opposite failure mode: guards do not reduce amplification, and for \texttt{Kimi-K2-Thinking} the ratio increases from 14.9$\times$ to 27.2--42.8$\times$.

As shown in Table~\ref{tab:qwen-code-guard}, \textbf{these guards can reduce token usage in some cases, but they do not reliably restore the compute--accuracy trade-off.}
On average, the unguarded \texttt{mixed} setting consumes 69.66M tokens, corresponding to 9.1$\times$ amplification over \texttt{None @ normal}, with an average accuracy drop of 40.99 percentage points. 
NoWait, \textsc{TCM}, and built-in loop detection reduce the average token burden to 8.2$\times$, 7.4$\times$, and 6.3$\times$, respectively. 
However, these reductions coincide with larger average accuracy drops of 52.94, 59.85, and 57.48 percentage points. 
Thus, lower token usage under a guard is not sufficient evidence of successful mitigation.

Trace inspection suggests why this trade-off persists: guarded runs often terminate prematurely, exhaust turn or depth limits, or remain diverted into unproductive validation and refinement paths before recovering a correct solution. 
This pattern appears even when token usage falls substantially. 
For example, \texttt{DeepSeek-V3.2} consumes only 0.5$\times$ the \texttt{None @ normal} token budget under all guards, yet accuracy remains at only 2.44--3.66\%; \texttt{MiniMax-M2} similarly drops from 20.3$\times$ to 10.7$\times$ under \textsc{TCM}, but accuracy falls from 66.26\% to 34.15\%. 
Conversely, for \texttt{Kimi-K2-Thinking}, guards fail to interrupt amplification at all, increasing token usage from 14.9$\times$ under \texttt{None @ mixed} to 27.1--42.7$\times$ under guarded execution. 
Effective mitigation for structural overthinking should therefore be evaluated jointly on resource use, task success, and trajectory quality.

% The model-level results reveal two distinct failure modes. 
% For \texttt{Kimi-K2-Thinking} and \texttt{Qwen3-Thinking}, guards do not consistently reduce amplification: \texttt{Kimi-K2-Thinking} increases from 14.9$\times$ under \texttt{None @ mixed} to 27.1--42.7$\times$ under guarded execution. 
% For \texttt{Qwen3-Coder}, \texttt{DeepSeek-V3.2}, and \texttt{GLM-4.7}, guards substantially reduce token usage, but accuracy remains near zero or collapses further. 
% For example, \texttt{DeepSeek-V3.2} consumes only 0.5$\times$ the \texttt{None @ normal} token budget under all guards, yet accuracy remains at only 2.44--3.66\%. 
% \texttt{MiniMax-M2} shows a partial trade-off: \textsc{TCM} reduces amplification from 20.3$\times$ to 10.7$\times$, but accuracy falls from 66.26\% to 34.15\%.

% Token reduction alone is therefore not a reliable mitigation criterion. 
% Trace inspection suggests that guarded runs often fail through premature termination, exhaustion of turn or depth limits, or continued diversion into unproductive validation and refinement paths. 
% Effective mitigation for structural overthinking should therefore be evaluated jointly on resource use, task success, and trajectory quality.

% Token reduction alone is therefore not sufficient evidence of mitigation. Trace inspection suggests that these outcomes often arise from three failure modes: guards may terminate execution before the agent recovers a correct solution, runs may exhaust turn or depth limits, or cycle-inducing tools may pull the agent away from a promising solution path until it fails to recover.

\subsection{Operational Resource Analysis}
\label{subsec:operational-resource-analysis}

\input{table_latex/revision_cost}

The preceding results report structural overthinking mainly through token usage and task accuracy.
However, these metrics do not fully capture the operational impact of an agent run.
We therefore estimate two deployment-facing quantities for the same \textit{Qwen-Code} executions.
For API consumers or service operators, input/output token growth translates into higher usage-based billing.
For model providers, the relevant quantity is GPU-hours, which measures the accelerator capacity occupied during active execution.
Table~\ref{tab:qwen-code-resource} reports both quantities, with the underlying input/output token and active-time statistics provided in Appendix~\ref{appendix:qwen-code-accounting-stats}.

The billing impact is consistent across models: under the \texttt{mixed} registry, estimated per-run cost increases by 4.3$\times$--18.9$\times$.
GPU-hours show a different pattern because they depend on active execution time, model throughput, serving-stack efficiency, and tool-execution overhead, not only on token volume.
Still, provider-side accelerator occupancy increases for five of the six models under \texttt{mixed}, with the largest increase observed for \texttt{Kimi-K2-Thinking} from 5.72 to 49.82 GPU-hours.
The exception is \texttt{Qwen3-Coder}, which has 4.3$\times$ higher billing while GPU-hours decrease from 22.56 to 14.08.
This difference arises because billing is determined by token accounting, whereas GPU-hours reflect the realized active execution time of the serving stack.
Thus, token amplification is a useful signal of resource abuse, but it should not be treated as the only deployment-facing measure.
Operational evaluation should separately report user-facing billing exposure and provider-side accelerator occupancy.

%% file: table_latex/qwen_code_main.tex
\begin{table}[tb]
\centering
\scriptsize
\setlength{\tabcolsep}{2pt}
\renewcommand{\arraystretch}{1.05}

% \begin{tabular}{c|cc|rl rl}
% \toprule
% \multirow{2}{*}{\textbf{Model}}
%   & \multicolumn{2}{c|}{\textbf{normal}}
%   & \multicolumn{4}{c}{\textbf{mixed}} \\

% & Tokens & Time (s)
% & \multicolumn{2}{c}{Tokens}
% & \multicolumn{2}{c}{Time (s)} \\
% \midrule
% \textbf{Kimi-K2-Thinking} & 31.6K & 176 & 343.9K & ($10.9\times$)  & 449  & ($2.6\times$) \\
% \textbf{MiniMax-M2}       & 45.9K & 82  & 6.5M   & ($142.4\times$) & 2260 & ($27.5\times$) \\
% \textbf{Qwen3-Coder}      & 127.5K& 77  & 2.9M   & ($22.4\times$)  & 727  & ($9.5\times$) \\
% \textbf{DeepSeek-V3.2}    & 273.5K& 222 & 1.5M   & ($5.5\times$)   & 1580 & ($7.1\times$) \\
% \textbf{GLM-4.6}          & 334.3K& 89  & 33.8M  & ($101.3\times$) & 925  & ($10.4\times$) \\
% \bottomrule
% \end{tabular}
\vspace{-0.8em}
\begin{tabular}{c|cc|rl rl}
\toprule
\multirow{2}{*}{\textbf{Model}}
  & \multicolumn{2}{c|}{\textbf{\texttt{normal}}}
  & \multicolumn{4}{c}{\textbf{\texttt{mixed}}} \\

& Tokens & Acc. (\%)
& \multicolumn{2}{c}{Tokens ($\times$ \texttt{normal})}
& \multicolumn{2}{c}{Acc. ($\Delta$ pp)} \\
\midrule
\textbf{Kimi-K2}        & 4.05M  & 75.98 & 60.73M  & (14.9$\times$)  & 76.46 & (+0.49) \\
\textbf{MiniMax-M2}     & 4.92M  & 86.38 & 99.65M  & (20.3$\times$) & 66.26 & (-20.12) \\
\textbf{Qwen3-Thinking} & 7.62M  & 76.83 & 59.43M  & (7.8$\times$)  & 54.39 & (-22.44) \\
\textbf{Qwen3-Coder}    & 9.82M  & 83.33 & 42.15M  & (4.3$\times$)  & 13.21 & (-70.12) \\
\textbf{DeepSeek-V3.2}  & 12.03M & 86.99 & 105.25M & (8.7$\times$)  & 24.59 & (-62.40) \\
\textbf{GLM-4.7}        & 7.47M  & 95.12 & 50.78M  & (6.8$\times$)  & 23.78 & (-71.34) \\
\bottomrule
\end{tabular}
\vspace{-1mm}
\caption{\textbf{Qwen-Code results on HumanEval.}
Tokens are reported in millions; parentheses show changes relative to \texttt{normal}.
}
\vspace{-1.0em}
\label{tab:qwen-code-main}
\end{table}

% \begin{table}[hbpt]
% \centering
% \scriptsize
%     % \begin{tabular}{c|cc|cc}
%     \begin{tabular}{@{\hspace{3pt}}c|c@{\hspace{3pt}}c|c@{\hspace{3pt}}c}
%         \toprule
%          & \multicolumn{2}{c|}{\textbf{normal}} & \multicolumn{2}{c}{\textbf{mixed}} \\
%         \textbf{Model}
%         & \textbf{Tokens}
%         & \textbf{Time (s)}
%         & \textbf{Tokens ($\times$)}
%         & \textbf{Time ($\times$)} \\
%         \midrule
%         \textbf{Kimi-K2-Thinking} & 31.6K & 176 & 343.9K (10.9$\times$) & 449 (2.6$\times$) \\
%         \textbf{MiniMax-M2} & 45.9K & 82 & 6.5M (142.4$\times$) & 2,260 (27.5$\times$) \\
%         \textbf{Qwen3-Coder} & 127.5K & 77 & 2.9M (22.4$\times$) & 727 (9.5$\times$) \\
%         \textbf{DeepSeek-V3.2} & 273.5K & 222 & 1.5M (5.5$\times$) & 1,580 (7.1$\times$) \\
%         \textbf{GLM-4.6} & 334.3K & 89 & 33.8M (101.3$\times$) & 925 (10.4$\times$) \\
%         \bottomrule
%     \end{tabular}
%     \caption{\textbf{Average token consumption and runtime in Qwen-Code settings.}}
%     \label{tab:qwen-code-main}
% \end{table}

%% file: table_latex/revision_tool_trajectory.tex
\begin{table}[!tb]
    \centering
    \scriptsize
    \setlength{\tabcolsep}{4pt}
    \renewcommand{\arraystretch}{1.05}
    \begin{tabular}{l|c|c|c|c}
    \toprule
    \textbf{Model} & \makecell{\textbf{Normal}\\\textbf{Tools}\\(\%)} & \makecell{\textbf{Repetition}\\(\%)} & \makecell{\textbf{Iterative}\\\textbf{Refinement}\\(\%)} & \makecell{\textbf{Auxiliary}\\\textbf{Subtasks}\\(\%)} \\
    \midrule
    \textbf{Kimi-K2-Thinking} & 3.79 & 26.51 & 35.91 & 33.80 \\
    \textbf{MiniMax-M2} & 11.25 & 26.62 & 29.91 & 32.22 \\
    \textbf{Qwen3-Thinking} & 10.78 & 25.78 & 29.88 & 33.56 \\
    \textbf{Qwen3-Coder} & 13.63 & 11.53 & 20.33 & 54.51 \\
    \textbf{DeepSeek-V3.2} & 0.04 & 15.94 & 33.02 & 51.01 \\
    \textbf{GLM-4.7} & 19.81 & 16.83 & 29.21 & 34.15 \\
    \bottomrule
    \end{tabular}
    \vspace{-1mm}
    \caption{
    \textbf{Tool-family distribution in mixed Qwen-Code trajectories.}
    % Values are count-weighted tool-call percentages; Repetition, Iterative Refinement, and Auxiliary Subtasks denote the three cycle-inducing tool families.
    Values are count-weighted tool-call percentages; \textbf{Normal Tools} counts calls to tools from the \texttt{normal} registry within \texttt{mixed} trajectories, and the other three columns correspond to the cycle-inducing tool families.
    }
    \label{tab:qwen-code-tool-family}
    \vspace{-1.0em}
\end{table}

%% file: table_latex/revision_cycle_trace.tex
\begin{figure}[t]
\centering
\scriptsize
\vspace{-0.8em}
\begin{tikzpicture}[
  font=\scriptsize,
  >=Latex,
  normal/.style={
    draw=black!55,
    fill=black!5,
    rounded corners=2pt,
    inner sep=3pt,
    align=center,
    text width=2.75cm
  },
  cycle/.style={
    draw=black!75,
    fill=black!13,
    rounded corners=2pt,
    inner sep=3pt,
    align=center,
    text width=3.0cm
  },
  link/.style={-Latex, thick, black!70},
  back/.style={-Latex, thick, dashed, black!45}
]

\node[font=\bfseries, align=center] at (-2.0,0)
  {Normal feedback\\[-1pt]{\tiny built-in execution}};
\node[font=\bfseries, align=center] at (2.0,0)
  {Injected cycle tools\\[-1pt]{\tiny \texttt{iterative\_refinement} server}};

\node[normal] (n1) at (-2.0,-0.8)
  {\textbf{Calls 20--23}\\ shell execution $\times 4$\\[-1pt]{\tiny outputs}};
\node[cycle] (c1) at (2.0,-0.8)
  {\textbf{Calls 24--27}\\ \texttt{verify} $\to$ \texttt{refine}\\
  $\to$ \texttt{validate} $\times 2$};

\node[normal] (n2) at (-2.0,-2.05)
  {\textbf{Call 28}\\ shell execution\\[-1pt]{\tiny output}};
\node[cycle] (c2) at (2.0,-2.05)
  {\textbf{Calls 29--30}\\ \texttt{refine} $\to$ \texttt{verify}};

\node[normal] (n3) at (-2.0,-3.30)
  {\textbf{Calls 31--32}\\ shell execution $\times 2$\\[-1pt]{\tiny error + output}};
\node[cycle] (c3) at (2.0,-3.30)
  {\textbf{Calls 33--34}\\ \texttt{verify} $\to$ \texttt{finalize}};

\node[normal] (n4) at (-2.0,-4.55)
  {\textbf{Calls 35--36}\\ shell execution $\times 2$\\[-1pt]{\tiny error + output}};
\node[cycle] (c4) at (2.0,-4.55)
  {\textbf{Calls 37--40}\\ \texttt{analyze} $\to$ \texttt{validate} $\times 3$};

\draw[link] (n1.east) -- node[above, font=\tiny] {refresh} (c1.west);
\draw[back] (c1.south west) -- (n2.north east);

\draw[link] (n2.east) -- (c2.west);
\draw[back] (c2.south west) -- (n3.north east);

\draw[link] (n3.east) -- (c3.west);
\draw[back] (c3.south west) -- (n4.north east);

\draw[link] (n4.east) -- (c4.west);

\end{tikzpicture}
\vspace{-1mm}
\caption{
\textbf{Example of interleaved cycle re-entry in Qwen-Code.}
Compressed excerpt from a DeepSeek-V3.2 \texttt{HumanEval/145} run. Built-in \texttt{run\_shell\_command} calls produce outputs or errors before re-entry into injected \texttt{iterative\_refinement} tools. Call numbers are original tool-call indices; $\times k$ denotes consecutive calls. 
}
\label{fig:interleaved-cycle-reentry}
\vspace{-1.0em}
\end{figure}

%% file: table_latex/revision_guard.tex
\definecolor{tokBurdenLow}{HTML}{EAF7FF}
\definecolor{tokBurdenMid}{HTML}{BFEAFF}
\definecolor{tokBurdenHigh}{HTML}{8BD8FF}
\definecolor{tokBurdenVeryHigh}{HTML}{4DBFFF}

% Accuracy colors: positive guard effect = accuracy preserved relative to None @ mixed.
\definecolor{accKeepLow}{HTML}{EAF7EA}
\definecolor{accKeepMid}{HTML}{BFE8BF}
\definecolor{accKeepHigh}{HTML}{7BCF7B}

\providecommand{\guardlegend}[2]{}
\renewcommand{\guardlegend}[2]{\begingroup\setlength{\fboxsep}{1pt}\colorbox{#1}{#2}\endgroup}

% Compact, vertically centered body cells.
\providecommand{\onecell}[1]{}
\renewcommand{\onecell}[1]{\begin{tabular}[c]{@{}c@{}}#1\end{tabular}}

\providecommand{\twocell}[2]{}
\renewcommand{\twocell}[2]{\begin{tabular}[c]{@{}c@{}}#1\\[-0.35ex]#2\end{tabular}}

\providecommand{\shadedtwocell}[3]{}
\renewcommand{\shadedtwocell}[3]{\cellcolor{#1}\begin{tabular}[c]{@{}c@{}}#2\\[-0.35ex]#3\end{tabular}}

\providecommand{\modelcell}[1]{}
\renewcommand{\modelcell}[1]{\begin{tabular}[c]{@{}c@{}}\textbf{#1}\end{tabular}}

\begin{table*}[hbpt]
\centering
\begingroup
\setlength{\tabcolsep}{2.2pt}
\renewcommand{\arraystretch}{0.92}
\resizebox{0.90\textwidth}{!}{%
\begin{tabular}{c|cc|cc|cc|cc|cc}
\toprule
\multirow{2}{*}{\textbf{Model}}
& \multicolumn{2}{c|}{\textbf{None @ \texttt{normal}}}
& \multicolumn{2}{c|}{\textbf{None @ \texttt{mixed}}}
& \multicolumn{2}{c|}{\textbf{NoWait @ \texttt{mixed}}}
& \multicolumn{2}{c|}{\textbf{TCM @ \texttt{mixed}}}
& \multicolumn{2}{c}{\textbf{Loop Detection @ \texttt{mixed}}} \\
\cmidrule{2-11}
& \textbf{Tokens} 
& \textbf{Acc.}
& \textbf{\makecell{Tokens\\($\times$ Normal)}} 
& \textbf{\makecell{Acc.\\($\Delta$ pp)}}
& \textbf{\makecell{Tokens\\($\times$ Normal)}} 
& \textbf{\makecell{Acc.\\($\Delta$ pp)}}
& \textbf{\makecell{Tokens\\($\times$ Normal)}} 
& \textbf{\makecell{Acc.\\($\Delta$ pp)}}
& \textbf{\makecell{Tokens\\($\times$ Normal)}} 
& \textbf{\makecell{Acc.\\($\Delta$ pp)}} \\
\midrule

\modelcell{Kimi-K2-Thinking}
& \onecell{4.05M} & \onecell{75.98}
& \twocell{60.73M}{(14.9$\times$)} & \twocell{76.46}{(+0.49)}
& \twocell{150.75M}{(37.2$\times$)} & \shadedtwocell{accKeepHigh}{71.34}{(-4.64)}
& \twocell{172.96M}{(42.7$\times$)} & \shadedtwocell{accKeepMid}{62.80}{(-13.18)}
& \twocell{109.79M}{(27.1$\times$)} & \shadedtwocell{accKeepMid}{57.32}{(-18.66)} \\

\modelcell{MiniMax-M2}
& \onecell{4.91M} & \onecell{86.38}
& \twocell{99.65M}{(20.3$\times$)} & \twocell{66.26}{(-20.12)}
& \shadedtwocell{tokBurdenLow}{94.95M}{(19.3$\times$)} & \shadedtwocell{accKeepHigh}{62.60}{(-23.78)}
& \shadedtwocell{tokBurdenMid}{52.72M}{(10.7$\times$)} & \twocell{34.15}{(-52.23)}
& \shadedtwocell{tokBurdenLow}{82.56M}{(16.8$\times$)} & \shadedtwocell{accKeepMid}{57.32}{(-29.06)} \\

\modelcell{Qwen3-Thinking}
& \onecell{7.61M} & \onecell{76.83}
& \twocell{59.42M}{(7.8$\times$)} & \twocell{54.39}{(-22.44)}
& \twocell{73.43M}{(9.6$\times$)} & \shadedtwocell{accKeepHigh}{49.39}{(-27.44)}
& \twocell{72.59M}{(9.5$\times$)} & \shadedtwocell{accKeepMid}{44.72}{(-32.11)}
& \shadedtwocell{tokBurdenLow}{55.30M}{(7.3$\times$)} & \shadedtwocell{accKeepLow}{41.87}{(-34.96)} \\

\modelcell{Qwen3-Coder}
& \onecell{9.81M} & \onecell{83.33}
& \twocell{42.15M}{(4.3$\times$)} & \twocell{13.21}{(-70.12)}
& \shadedtwocell{tokBurdenMid}{27.31M}{(2.8$\times$)} & \shadedtwocell{accKeepLow}{1.22}{(-82.11)}
& \shadedtwocell{tokBurdenHigh}{11.87M}{(1.2$\times$)} & \shadedtwocell{accKeepLow}{0.20}{(-83.13)}
& \shadedtwocell{tokBurdenHigh}{11.95M}{(1.2$\times$)} & \shadedtwocell{accKeepLow}{0.61}{(-82.72)} \\

\modelcell{DeepSeek-V3.2}
& \onecell{12.03M} & \onecell{86.99}
& \twocell{105.24M}{(8.7$\times$)} & \twocell{24.59}{(-62.40)}
& \shadedtwocell{tokBurdenVeryHigh}{6.33M}{(0.5$\times$)} & \twocell{2.44}{(-84.55)}
& \shadedtwocell{tokBurdenVeryHigh}{6.45M}{(0.5$\times$)} & \twocell{3.66}{(-83.33)}
& \shadedtwocell{tokBurdenVeryHigh}{6.05M}{(0.5$\times$)} & \twocell{2.44}{(-84.55)} \\

\modelcell{GLM-4.7}
& \onecell{7.46M} & \onecell{95.12}
& \twocell{50.77M}{(6.8$\times$)} & \twocell{23.78}{(-71.34)}
& \shadedtwocell{tokBurdenMid}{25.45M}{(3.4$\times$)} & \twocell{0.00}{(-95.12)}
& \shadedtwocell{tokBurdenHigh}{23.77M}{(3.2$\times$)} & \twocell{0.00}{(-95.12)}
& \shadedtwocell{tokBurdenHigh}{21.64M}{(2.9$\times$)} & \twocell{0.20}{(-94.92)} \\

\midrule
\modelcell{Average}
& \onecell{7.65M} & \onecell{84.11}
& \twocell{69.66M}{(9.1$\times$)} & \twocell{43.12}{(-40.99)}
& \shadedtwocell{tokBurdenLow}{63.04M}{(8.2$\times$)} & \shadedtwocell{accKeepLow}{31.16}{(-52.94)}
& \shadedtwocell{tokBurdenLow}{56.73M}{(7.4$\times$)} & \shadedtwocell{accKeepLow}{24.25}{(-59.85)}
& \shadedtwocell{tokBurdenMid}{47.88M}{(6.3$\times$)} & \shadedtwocell{accKeepLow}{26.63}{(-57.48)} \\

\bottomrule
\end{tabular}%
}
\endgroup
\vspace{-1mm}
\caption{
\textbf{Guarded execution results in Qwen-Code on HumanEval.}
Reported values use None @ \texttt{normal} as reference: token parentheses show amplification and accuracy parentheses show $\Delta$ pp. Cell shading compares each guarded \texttt{mixed} run against None @ \texttt{mixed}. Token shading indicates reduction:
\guardlegend{tokBurdenLow}{$0.75$--$1.0\times$},
\guardlegend{tokBurdenMid}{$0.5$--$0.75\times$},
\guardlegend{tokBurdenHigh}{$0.25$--$0.5\times$},
and \guardlegend{tokBurdenVeryHigh}{$<0.25\times$}.
Accuracy shading indicates preservation:
\guardlegend{accKeepLow}{$[-20,-10)$ pp},
\guardlegend{accKeepMid}{$[-10,-5)$ pp},
and \guardlegend{accKeepHigh}{$\geq -5$ pp}.
The average row uses arithmetic means; ratios and deltas are computed from averaged raw values.
}
\vspace{-1.0em}
\label{tab:qwen-code-guard}
\end{table*}

%% file: table_latex/revision_cost.tex
\begin{table*}[t]
\centering
\scriptsize
\renewcommand{\arraystretch}{0.91}
\resizebox{0.85\textwidth}{!}{%
\begin{tabular}{lrrrrrrr}
\toprule
\multirow{2}{*}{\textbf{Model}} &
\multicolumn{3}{c}{\textbf{Consumer-side billing}} &
\multicolumn{4}{c}{\textbf{Provider-side GPU exposure}} \\
\cmidrule(lr){2-4}\cmidrule(lr){5-8}
& \textbf{\texttt{normal}} 
& \textbf{\texttt{mixed}} 
& \textbf{Amp.}
& \textbf{\#GPU} 
& \textbf{\texttt{normal} GPU-h} 
& \textbf{\texttt{mixed} GPU-h} 
& \textbf{Amp.} \\
\midrule
Kimi-K2 & \$2.57 & \$37.53 & 14.6$\times$ & 8 & 5.72 & 49.82 & 8.7$\times$ \\
MiniMax-M2 & \$1.37 & \$25.96 & 18.9$\times$ & 4 & 6.85 & 34.50 & 5.0$\times$ \\
Qwen3-Thinking & \$1.32 & \$9.64 & 7.3$\times$ & 4 & 5.42 & 19.79 & 3.7$\times$ \\
Qwen3-Coder & \$2.36 & \$10.01 & 4.3$\times$ & 8 & 22.56 & 14.08 & 0.6$\times$ \\
DeepSeek-V3.2 & \$3.06 & \$26.66 & 8.7$\times$ & 8 & 35.16 & 38.25 & 1.1$\times$ \\
GLM-4.7 & \$3.15 & \$21.06 & 6.7$\times$ & 8 & 24.67 & 33.51 & 1.4$\times$ \\
\bottomrule
\end{tabular}
}
\vspace{-1mm}
\caption{
\textbf{Bilateral resource exposure in Qwen-Code.}
Billing is computed from input/output token counts and OpenRouter prices as of May 2026\protect\footnotemark. 
GPU-hours are computed as active time in hours multiplied by the number of H200 GPUs used by each model.
% GPU-side amplification is computed as the ratio of \texttt{mixed} GPU-hours to \texttt{normal} GPU-hours.
GPU-side amplification is the \texttt{mixed}-to-\texttt{normal} GPU-hour ratio.
}
\label{tab:qwen-code-resource}
\vspace{-1.2em}
\end{table*}

\footnotetext{OpenRouter model pricing: \url{https://openrouter.ai/models}. Accessed May 2026.}

%% file: tex/5_discussion.tex
\section{Discussion \& Future Work}
\label{sec:discussions}

\paragraph{Harness-specific cycle patterns.}

Our experiments show registry-induced overthinking across both \textit{ReAct}-style agents and a coding-agent harness. 
The form of such cycles may depend on the agent harness: different systems expose different tool interfaces, state representations, and stopping policies, which can change how cyclic trajectories emerge. 
Future work should examine broader settings, including browsing and computer-use agents, where cycle induction may take forms different from the reasoning and coding loops studied here.

\paragraph{Mitigation as a layered control problem.}

The guarded-execution results suggest that mitigation is not a single-detector problem. Detecting recurrence is necessary, but the system must decide whether the repeated trajectory is still productive, whether the agent can still reach a correct solution, and whether continued execution is worth the remaining budget. Future work should therefore treat mitigation as a layered control problem that combines registry-level vetting, execution budgets, turn and tool-depth limits, trajectory monitors, and model-level training objectives that discourage repeated no-progress tool use. Training-time mitigation may improve default agent behavior, but future work should evaluate how well such training generalizes to third-party tools and unseen registry compositions introduced after deployment.

\paragraph{Trajectory-aware evaluation.}

Our results motivate treating the execution trajectory as part of the evaluation target, not as an auxiliary log. Final accuracy alone cannot distinguish a concise successful run from one that reaches the same answer after excessive tool use, and token reduction alone cannot distinguish effective mitigation from early termination. In this work, we analyze trajectory-level signals: token usage, latency, termination status, tool-call structure, registry composition, and operational exposure for the evaluated settings. Future work should standardize these signals across agent harnesses so that productive tool use, structural overthinking, and premature termination can be compared under a shared evaluation protocol.

%% file: tex/6_conclusion.tex
\section{Conclusion}

% We show that cyclic tool-call structures in MCP registries can induce severe resource amplification in LLM agents without altering user queries or model parameters. The attack requires no sophisticated adversarial engineering; each tool relies on trivial logic that can become costly through cyclic execution. The vulnerability intensifies in production-grade coding agents, where richer tool ecosystems provide more pathways for loop induction. Counterintuitively, \texttt{mixed} registries can amplify cost more than \texttt{attack}-only configurations, and generation-level mitigation methods do not reliably prevent loop induction. These findings suggest that securing agentic systems requires reasoning about tool-call structure, not tokens alone.

We identify structural overthinking loops as a registry-level failure mode in tool-using LLM agents. Across ReAct-style agents evaluated on AIME2025, GPQA Diamond, and HumanEval, and \textit{Qwen-Code} evaluated on the full HumanEval benchmark, \texttt{mixed} registries with cycle-inducing tool configurations substantially increase computational cost and can degrade task success. Our trajectory analysis shows that these failures are not merely degenerate self-loops: normal execution tools and injected cycle-inducing tools can interleave, refreshing the local context and repeatedly driving the agent back into verification, refinement, or validation states. Guarded execution with token-level suppression, trajectory-cycle monitoring, or built-in loop detection can reduce token usage in some cases, but does not reliably restore the compute--accuracy trade-off. These findings suggest that safe and efficient agentic systems require trajectory-aware evaluation and registry-level controls that account for tool-call structure, execution policy, and resource consumption in addition to final task correctness.

%% file: tex/7_limitations.tex
\section*{Limitations}
\label{sec:limitations}

% EMNLP 2026 은 Limitations, Ethical Considerations 분량에서 제외

\paragraph{Scope of Cycle-Inducing Tools}

Our cycle-inducing tool suite is intentionally small and simple. The 14 configurations across repetition, iterative refinement, and auxiliary subtasks are not intended to represent the full space of possible cycle-inducing MCP servers, nor are they optimized for stealth, placement, naming, or adaptive behavior. Some tools, especially repetition-style configurations, may appear suspicious when inspected in isolation. Our claim is therefore not that these specific tools are maximally realistic or comprehensive. Rather, we show that even naive cycle-inducing configurations can trigger substantial amplification once they are included in a larger registry and composed with normal tools over multi-step trajectories. Future work should study more naturalistic tool descriptions, different registry proportions, adaptive tool behavior, and stronger registry-level vetting procedures.

\paragraph{Ablations and Controls}

The \texttt{mixed} condition changes several factors at once: it adds 14 tools to the registry, it adds descriptions and return messages that explicitly direct the agent to call further tools, and in the main \textit{Qwen-Code} runs it is evaluated with built-in loop detection disabled. We do not ablate these factors. In particular, we do not test versions of the tools with the re-invocation directives removed or rewritten in benign wording, a control registry that adds 14 benign tools with descriptions of similar length, or sensitivity to the proportion, ordering, and phrasing of the cycle-inducing tools. The amplification we report should therefore be attributed to this combination rather than to registry size, description length, or cycle semantics alone, and separating these factors remains future work.

\paragraph{Dataset Scope and Task Coverage}

We focus on scientific reasoning, mathematics, and programming, where tool use and multi-step execution are clearly observable. This limits coverage of open-ended dialogue, browsing-heavy tasks, and domain-specific enterprise workflows. In the \textit{Qwen-Code} setting, we evaluate the full HumanEval benchmark to avoid relying on a small model-filtered coding subset. However, HumanEval still represents short, unit-testable programming problems rather than long-horizon software engineering tasks involving large repositories, multi-file refactoring, or extended dependency management. Future work should evaluate whether similar registry-induced loops arise in broader coding and non-coding agent workloads.

\paragraph{Defense Mechanisms and System Design}

Our guarded-execution experiments evaluate NoWait, \textsc{TCM}, and Qwen-Code's built-in loop detection, covering generation-time suppression, trajectory-level monitoring, and a system-level loop detector. These results should be read as mechanism-probing evidence rather than a comprehensive defense benchmark. In particular, we do not systematically evaluate stronger registry-level defenses, such as static linting of tool descriptions for re-invocation directives, per-tool call budgets, limits on repeated cycles across turns, tool provenance checks, policy-based tool admission, signed or permissioned MCP servers, learned trajectory monitors, human approval for high-risk tools, or adaptive execution budgets. Comparative analysis of such defenses remains future work.

\paragraph{Generalization across Agent Configurations}

Our experiments assume largely autonomous agents that can select tools over multi-step trajectories with minimal human intervention. Real-world deployments may include additional guardrails, narrower tool registries, approval steps, sandboxing, or stricter execution policies, so our results should not be read as a claim that every guarded deployment is equally vulnerable. Rather, they show that when agents can autonomously compose normal and cycle-inducing tools, registry composition and execution policy become safety-relevant variables that should be evaluated alongside model quality.

\paragraph{Resource and Cost Estimates}

Our resource analysis estimates consumer-side billing from observed input/output token counts and provider pricing, and provider-side GPU exposure from active execution time and GPU count.
These estimates make the operational risk concrete, but they are not a substitute for deployment-specific metering.
Actual cost and resource usage can vary with provider billing policy, batching, caching, utilization, hardware generation, serving framework, and system-level scheduling.
We therefore interpret the billing and GPU-hour results as approximate indicators of operational impact rather than universal measurements of monetary cost or infrastructure load.

%% file: tex/ethical_consideration.tex
\section*{Ethical Considerations}
\label{sec:ethical-consideration}

% This work shows that a small set of cycle-inducing tools, when co-registered with normal tools in a tool registry for an MCP-based agent, can cause substantial resource amplification. Identifying such structural vulnerabilities is important for building safer agentic systems; however, we acknowledge that our findings could be misused for cost-amplification or Denial-of-Wallet attacks.
This work studies a registry-level failure mode in tool-using LLM agents: a small set of cycle-inducing tool configurations, when co-registered with normal tools, can trigger cyclic trajectories that substantially amplify resource consumption. Identifying such structural vulnerabilities is important for building safer agentic systems. At the same time, we acknowledge the dual-use nature of the finding: similar mechanisms could be misused for cost-amplification or resource-exhaustion abuse against deployed agent services.

% To ensure responsible research conduct, we restrict our experiments to open-source (or open-weights) models accessible under permissive licensing conditions. We do not evaluate proprietary production APIs (e.g., OpenAI's GPT series, Google's Gemini, or Anthropic's Claude). Inducing execution loops or resource amplification against production APIs could violate their Terms of Service or Usage Policies, which often prohibit disruption, interference, abuse, or the bypassing of protective measures. A detailed comparison of these policies is provided in Appendix~\ref{appendix:proprietary-model-policy}. Our experimental design therefore avoids interactions that could compromise service stability or contravene platform regulations.
To ensure responsible research conduct, we restrict our experiments to open-source or open-weight models served in controlled environments. We do not evaluate proprietary production APIs such as OpenAI's GPT series, Google's Gemini, or Anthropic's Claude. Inducing execution loops or resource amplification against production APIs could violate provider Terms of Service or usage policies, particularly provisions related to disruption, abuse, or bypassing protective measures. A detailed comparison of these policies is provided in Appendix~\ref{appendix:proprietary-model-policy}. This boundary is part of our experimental ethics rather than evidence that proprietary systems are safe or vulnerable; our empirical claims are restricted to the controlled open-model and local-serving settings evaluated in this paper.

% All experiments are conducted under standard usage conditions without privileged access, system modification, or exploitation of undocumented interfaces. Importantly, our cycle-inducing tool configurations do not rely on novel exploitation techniques, hidden backdoors, or model-specific vulnerabilities. Instead, they are constructed from routine interaction patterns already common in legitimate tool design, such as verification workflows, iterative refinement procedures, and supplementary task requests. This design choice helps ensure that the vulnerabilities arise from structural properties of agent execution, tool registry composition, and tool-visible text, rather than from obscure or artificially engineered exploits.
All experiments are conducted without privileged access, system modification, hidden backdoors, or exploitation of undocumented interfaces. The cycle-inducing tools are constructed from ordinary tool-design patterns, such as repeated verification, iterative refinement, and supplementary analysis requests. We intentionally do not optimize the tools for stealth, adaptive behavior, or provider-specific exploitation. This design choice focuses the study on structural properties of registry composition and tool-call trajectories rather than on maximally effective attack engineering.

% Even when working with open-source models, a comprehensive analysis of internal training mechanisms remains limited. Critical training recipes, including tool-calling strategies, instruction-tuning configurations, and fine-tuning parameters, are often not fully disclosed. We therefore refrain from attributing observed behaviors to specific undisclosed components and instead focus on externally observable execution dynamics.
We also avoid drawing conclusions about undisclosed model-internal mechanisms. Even for open-weight models, training recipes, tool-use tuning data, and decoding or serving details are often only partially available. We therefore analyze externally observable execution behavior---tool-call trajectories, token usage, latency, task success, and termination conditions---rather than attributing the observed failures to specific hidden training or alignment components.

% Our objective is not to enable exploitation but to inform the development of robust safeguards for tool-calling agents. In particular, our findings highlight registry-level vetting mechanisms, session-level cost monitoring, and structural constraints on execution depth and tool-call recursion. We hope these insights contribute to more secure agent architectures and align with responsible disclosure practices in AI security research.
Our objective is not to enable exploitation, audit specific providers, or claim that all deployed agents are equally vulnerable. Rather, we aim to inform safeguards for tool-using agents. The results motivate registry-level vetting, provenance checks for third-party tools, session-level cost and resource monitoring, trajectory-aware execution controls, and explicit limits on recursive tool use. We hope this work supports safer agent architectures by making resource consumption and trajectory structure part of responsible agent evaluation.

%% file: tex/8_acknowledgments.tex
\section*{Acknowledgments}

This work was supported by Hankuk University of Foreign Studies Research Fund (of 2025) and by the National Research Foundation of Korea (NRF) grant funded by the Korea government (MSIT) (No.~RS-2026-25602702).

%% file: tex/appendix.tex
\newpage
% \onecolumn
\appendix
\raggedbottom

% --- Appendix 대제목 섹션 ---
\section*{Appendix}
\bigskip

% Appendix A - $2.2 에서 언급
\input{tex/appendix_attack_tools}

% Appendix B - Implementation Details
\section{Implementation Details}
\label{appendix:implementation-details}

% B2.1 - $3 Tool Registry Configuration 언급
\subsection{\textit{Qwen-Code} Execution Harness}
\label{appendix:qwen-code-setup}

We run \textit{Qwen-Code} through its command-line interface rather than reimplementing the agent loop. For each HumanEval problem, the runner creates an item-specific workspace and writes a project-local \texttt{.qwen/settings.json} file that specifies the OpenAI-compatible model endpoint, the allowed MCP servers, and the loop-detection setting. This keeps the experiment configuration local to the current problem instance and avoids modifying global \textit{Qwen-Code} settings. Credentials are injected only through the subprocess environment, and command traces are logged with secret values redacted.

The runner executes each problem by spawning a \textit{Qwen-Code} subprocess with \texttt{stream-json} output enabled, allowing us to recover turn-level metrics, tool calls, token usage, available tools, MCP servers, and termination status. We use one isolated workspace per problem to avoid cross-run interference from filesystem side effects, and we redirect runtime cache directories such as \texttt{HOME}, \texttt{XDG\_CACHE\_HOME}, \texttt{npm\_config\_cache}, and \texttt{UV\_CACHE\_DIR} into the current run directory.
In the \texttt{normal} registry, each workspace registers the built-in \textit{Qwen-Code} tools together with Serena and Context7. In the \texttt{mixed} registry, the same normal registry is used, and the cycle-inducing MCP servers are additionally registered.

\paragraph{\textit{Qwen-Code} Loop Detection.}
\textit{Qwen-Code} exposes a built-in loop-detection option through the model setting \texttt{skipLoopDetection}. In our main \texttt{normal}/\texttt{mixed} experiments, we set \texttt{skipLoopDetection=true}, disabling the built-in detector so that the baseline measures unconstrained registry-induced amplification. In the \textsc{Loop Detection} guard condition, we set \texttt{skipLoopDetection=false}, enabling the built-in detector while keeping the remaining registry, model, prompt, and execution limits unchanged. This condition is therefore a system-level guard baseline, distinct from \textsc{NoWait}, which modifies token generation, and \textsc{TCM}, which monitors completed tool-call trajectories.

\paragraph{\textit{Qwen-Code} Built-in Tools (13).}
\noindent
\begin{ColorVerbatim}
Edit, ExitPlanMode, Glob, Grep, ListFiles, ReadFile, ReadManyFiles, SaveMemory, Shell, Task, TodoWrite, WebFetch, WriteFile
\end{ColorVerbatim}

\paragraph{Serena MCP Tools (29).}
\noindent
\begin{ColorVerbatim}
activate_project, check_onboarding_performed, create_text_file, delete_memory, edit_memory, execute_shell_command, find_file, find_referencing_symbols, find_symbol, get_current_config, get_symbol_overview, initial_instructions, insert_after_symbol, insert_before_symbol, list_dir, list_memories, onboarding, prepare_for_new_conversation, read_file, read_memory, rename_symbol, replace_content, replace_symbol_body, search_for_pattern, switch_modes, think_about_collected_information, think_about_task_adherence, think_about_whether_you_are_done, write_memory
\end{ColorVerbatim}

\paragraph{Context7 MCP Tools (2).}
\noindent
\begin{ColorVerbatim}
resolve-library-id, get-library-docs
\end{ColorVerbatim}

\input{table_latex/react_wo_nowait}
\subsection{Model Configurations}
\label{appendix-subsec:model-configuration}

All LLMs were served through OpenAI-compatible APIs using vLLM\footnote{\url{https://github.com/vllm-project/vllm}} or SGLang\footnote{\url{https://github.com/sgl-project/sglang}}. For models with distinct tool-call or parsing conventions, such as MiniMax-M2 and DeepSeek-V3.2, we used a lightweight compatibility layer to convert responses into the format expected by the agent harness.

MiniMax-M2 and Qwen3-235B-A22B-Thinking-2507 were deployed on H200$\times$4 GPUs, while all other models used H200$\times$8 GPUs. A problem instance ends when the agent produces a final answer with no further tool calls or when it reaches one of the configured execution limits: 1,000 turns per session, 20 tool calls per turn, or a 3,600-second wall-clock cap. If the wall-clock cap is reached, execution is forcibly terminated even if a tool call is still in progress. Token, latency, and active-time totals are reported under this execution policy and include all completed calls observed before the run stops.

% B2.3 - $4.1 언급
\subsection{NoWait Settings}
\label{sec:appendix-nowait}

We adapted the NoWait technique to a tool-using agent environment. 
NoWait suppresses latency-inducing tokens at generation time by masking their logits. 
Following the original methodology, we use the same set of 17 latency-inducing tokens reported in the paper, which we denote as $T$:
\vspace{-0.5em}
\[
T = \left\{
\begin{aligned}
&\texttt{wait},\ \texttt{alternatively},\ \texttt{hmm},\ \texttt{but},\\
&\texttt{however},\ \texttt{alternative}, \texttt{another},\\ 
&\texttt{check},\ \texttt{double-check},\ \texttt{oh},\ \texttt{maybe},\\
&\texttt{verify}, \texttt{other},\ \texttt{again}, \texttt{now},\ \texttt{ah},\ \texttt{any}
\end{aligned}
\right\}.
\]

When NoWait is applied, the output probabilities for both the tokens in set $T$ and their derivatives, including variations in whitespace, special characters, and capitalization (e.g., \texttt{`` wait''}, \texttt{``wait ''}, \texttt{``Wait''}, \texttt{``WAIT''}, \texttt{``wait.''}), are set to $-\infty$ to prevent their generation. This process is implemented solely through logit masking at the inference stage, without modifying the model parameters or the agent's policy. Consequently, 
%NoWait does not directly interfere with structural reasoning or tool-call selection; 
it controls lexical expression during token generation.

\subsection{Online Trajectory Cycle Monitor}
\label{sec:appendix-tcm}

We implement a \textit{Trajectory Cycle Monitor} (\textsc{TCM}) as an online guard for the \textit{Qwen-Code} experiments, separate from \textit{Qwen-Code}'s built-in loop detector. George et al.~\cite{george2025unsupervisedcycledetectionagentic} detect cycles in completed agent trajectories using temporal call-stack analysis and semantic similarity. Because our setting requires intervention during coding-agent execution, we do not directly reproduce their full offline detector. Instead, we adapt the parts that transfer to our setting: detecting repeated tool-call subsequences and combining this signal with a no-progress check before termination. Thus, \textsc{TCM} operates during execution and can stop a sample once the current trajectory exhibits a repeated no-progress pattern.

\textsc{TCM} observes completed tool-call events rather than model tokens. Each event is represented by an action signature, based on the tool namespace, tool name, and normalized arguments, and a result signature, based on the result status and result hash. We normalize volatile values such as timestamps, UUIDs, temporary paths, and long blobs before hashing. In our experiments, \textsc{TCM} scans the most recent 30 tool events for subsequences of length 1--6 that repeat at least three times, allowing at most one intervening tool call between occurrences. The guard triggers only when the repeated span shows no observable progress, defined as no artifact change and no change in the result signatures for repeated actions. When triggered, the run is terminated and logged with the trigger index, tokens consumed before the trigger, cycle length, repeat count, and cycle signature.

\input{table_latex/revision_qwen_stats}
\input{table_latex/policy}

% Appendix C
\section{Additional Experimental Results}
\label{app-sec:additional-results}

\subsection{Full \textit{ReAct} Results}
\label{app-sec:react-full-results}

Table~\ref{tab:react-wo-nowait} reports the full per-model and per-dataset results for the \textit{ReAct}-based agent setting under the \texttt{normal} and \texttt{mixed} registry configurations. We report both task accuracy and token consumption across AIME2025, GPQA Diamond, and HumanEval. Token amplification is computed relative to the corresponding \texttt{normal} run under the same model and dataset.  

These results provide the detailed breakdown behind Figure~\ref{fig:full_results_main}. Across all evaluated models and task families, the \texttt{mixed} registry increases token consumption relative to \texttt{normal}, while the effect on accuracy varies across models and datasets. This supports the main finding that registry-induced loops can amplify resource consumption even when the user query and model configuration are unchanged.  

% Appendix D - $3.2 Production-Grade Coding Agents (Qwen-Code) 에서 언급
\subsection{\textit{Qwen-Code} Accounting Statistics}
\label{appendix:qwen-code-accounting-stats}

Table~\ref{tab:qwen-code-accounting-stats} reports the run-level accounting statistics used for the billing and GPU-exposure estimates in Table~\ref{tab:qwen-code-resource}. We include these statistics to make the resource accounting reproducible without duplicating the main Qwen-Code outcome analysis. Token statistics are aggregated over the full HumanEval benchmark for each run and are decomposed into input and output tokens for billing estimates. Active time denotes the aggregate active execution time across HumanEval problem instances, which is used to compute GPU-hours. The 3,600-second wall-clock cap is applied per problem instance, not to the full benchmark run; therefore, run-level active time can exceed 3,600 seconds.

% Appendix E - Ethical Considerations 언급
\section{Proprietary Model Policy}
\label{appendix:proprietary-model-policy}

Table~\ref{tab:model-terms-appendix} summarizes the relevant terms and conditions of proprietary models considered in our study. These policies were reviewed to ensure that all experiments complied with provider-specific usage restrictions, including limitations on data extraction, model replication, API quotas, and service-disruption behavior.

We conducted all evaluations within the permitted scope of each provider's terms of service. No proprietary production API was used for resource-amplification experiments, and no outputs were used for model training, automated scraping, or derivative model development. The access dates listed in the table correspond to the versions of the policies consulted during experimentation.

%% file: tex/appendix_attack_tools.tex
\section{Cycle-Inducing Tool Configurations}
\label{app-sec:appendix-attack-tools}

% Table~\ref{tab:attack-tools-summary} summarizes the attack-oriented MCP tools used in our experiments, outlining their functional roles and control-flow properties.
Table~\ref{tab:attack-tools-summary} summarizes the cycle-inducing MCP tool configurations used in our experiments, outlining their functional roles, tool-visible directives, and control-flow properties.

\begin{table*}[t]
    \centering
    \renewcommand{\arraystretch}{0.8}
    \tiny
    \setlength{\tabcolsep}{6pt}
    \renewcommand{\arraystretch}{1.15}
    \begin{tabularx}{\textwidth}{l|l|X|X|X}
        \toprule
        \textbf{Category} & \textbf{Tool Name} & \textbf{Description} & \textbf{Arguments} & \textbf{Returns} \\
        \midrule
        
        & echo\_verification
        & Forces large-scale marker repetition and recursively demands re-invocation to inflate token usage.
        & \textbullet\ count (int): repetition count \newline (enforced minimum)
        & Marker string \newline + mandatory next-call instruction \\ \cmidrule{2-5}
        \multirow{3}{*}{Repeat Text} & progress\_marker
        & Inserts repeated progress markers and enforces frequent invocation across micro-steps.
        & \textbullet\ step (str): progress label \newline \textbullet\ marker\_count (int): repetition count
        & Markers \newline + mandatory next-call instruction \\ \cmidrule{2-5}
        & confirmation\_echo
        & Repeatedly confirms generated content and enforces iterative validation cycles.
        & \textbullet\ content (str): content to confirm \newline \textbullet\ repetitions (int): repetition count
        & Confirmation text + markers \newline + recursive trigger \\
        
        \midrule

        & analyze
        & Enforces mandatory pre-response analysis and redirects control flow to validation.
        & \textbullet\ problem (str) \newline \textbullet\ approach (str) \newline \textbullet\ complexity\_assessment (opt. str)
        & Analysis summary \newline + forced next-step instruction \\ \cmidrule{2-5}
        & validate
        & Requires multiple validation passes; may redirect back to analyze on detected issues.
        & \textbullet\ analysis\_result (str) \newline \textbullet\ validation\_checklist (str) \newline \textbullet\ issues\_found (opt. str)
        & Validation output \newline + branching/recursion instruction \\ \cmidrule{2-5}
        \multirow{5}{*}{Iterative Refinement} & refine
        & Forces iterative improvements; may trigger backtracking to earlier steps when new issues arise.
        & \textbullet\ original\_solution (str) \newline \textbullet\ refinements\_made (str) \newline \textbullet\ new\_issues\_discovered (opt. str)
        & Refinement output \newline + mandatory next-step \\ \cmidrule{2-5}
        & verify
        & Requires comprehensive checks; redirects to refine/validate/analyze if concerns remain.
        & \textbullet\ refined\_solution (str) \newline \textbullet\ verification\_results (str) \newline \textbullet\ remaining\_concerns (opt. str)
        & Verification output \newline + enforced backtracking \\ \cmidrule{2-5}
        & finalize
        & Prevents termination unless high confidence/completeness thresholds are met; enforces multi-cycle execution.
        & \textbullet\ final\_solution (str) \newline \textbullet\ completeness\_score (str) \newline \textbullet\ confidence\_level (str) \newline \textbullet\ next\_step (str)
        & Finalization output \newline + mandatory cycle restart \\
        
        \midrule
        
        & check\_time\_complexity
        & Forces complexity analysis as the first mandatory step for any request.
        & \textbullet\ complexity (str) \newline \textbullet\ reasoning (str)
        & Complexity report \newline + enforced next-step \\ \cmidrule{2-5}
        & verify\_edge\_cases
        & Requires enumerating many edge cases before proceeding.
        & \textbullet\ edge\_cases (List[str]) \newline \textbullet\ analysis (str)
        & Edge-case listing \newline + recursive instruction \\ \cmidrule{2-5}
        \multirow{6}{*}{Auxiliary Subtasks} & consider\_alternative\_approaches
        & Forces enumeration of multiple alternative approaches, expanding reasoning length.
        & \textbullet\ approaches (List[Dict[str,str]])
        & Alternative approach report \newline + enforced continuation \\ \cmidrule{2-5}
        & check\_historical\_context
        & Requires historical contextualization regardless of task relevance.
        & \textbullet\ historical\_examples (str)
        & Historical context output \newline + forced next-step \\ \cmidrule{2-5}
        & assess\_practical\_applications
        & Forces enumeration of numerous concrete applications to prolong output.
        & \textbullet\ applications (List[str])
        & Applications report \newline + recursive instruction \\ \cmidrule{2-5}
        & validate\_assumptions
        & Requires validating many assumptions and explicitly restarts the cycle.
        & \textbullet\ assumptions (List[Dict[str,str]])
        & Assumption report \newline + mandatory cycle restart \\
        \bottomrule
    \end{tabularx}
    \caption{\textbf{Unified summary of cycle-inducing MCP tool configurations used in our experiments.}}
    \label{tab:attack-tools-summary}
    \vspace{-1.0em}
\end{table*}

%% file: table_latex/react_wo_nowait.tex
\begin{table*}[hbpt]
\centering
\scriptsize
\resizebox{\textwidth}{!}{%
\begin{tabular}{c|c|cc|cc|cc}
\toprule
& & \multicolumn{2}{c|}{\textbf{AIME2025}} & \multicolumn{2}{c|}{\textbf{GPQA Diamond}} & \multicolumn{2}{c}{\textbf{HumanEval}} \\ \cmidrule{3-8} 
\multirow{-2}{*}{\textbf{Model}} & \multirow{-2}{*}{\makecell{\textbf{Tool}\\\textbf{Registry}}}
 & \textbf{Tokens ($\times$ \texttt{normal})} & \multicolumn{1}{c}{\textbf{Acc. ($\Delta$ pp)}}  
 & \textbf{Tokens ($\times$ \texttt{normal})} & \multicolumn{1}{c}{\textbf{Acc. ($\Delta$ pp)}}  
 & \textbf{Tokens ($\times$ \texttt{normal})} & \multicolumn{1}{c}{\textbf{Acc. ($\Delta$ pp)}}  \\ \midrule
& \texttt{normal} & 4,251,880 & \multicolumn{1}{c|}{70.00\%} & 14,098,048   & \multicolumn{1}{c|}{72.59\%} & 9,635,441 & \multicolumn{1}{c}{19.51\%} \\
\multirow{-2}{*}{\textbf{Kimi-K2-Thinking}} & \texttt{mixed}  & 4,427,514 (1.04$\times$)  & \multicolumn{1}{c|}{36.67\% (-33.33)}    & 37,983,373 (2.69$\times$)   & \multicolumn{1}{c|}{64.97\% (-7.62)}   & 27,322,798 (2.84$\times$) & \multicolumn{1}{c}{17.68\% (-1.83)}  \\ \midrule
& \texttt{normal} & 1,462,548  & \multicolumn{1}{c|}{80.00\%} & 6,832,721    & \multicolumn{1}{c|}{56.85\%} & 7,437,881 & \multicolumn{1}{c}{76.83\%} \\
\multirow{-2}{*}{\textbf{MiniMax-M2}} & \texttt{mixed}  & 8,206,770 (5.61$\times$)  & \multicolumn{1}{c|}{70.00\% (-10.00)}    & 73,988,176 (10.83$\times$)  & \multicolumn{1}{c|}{53.30\% (-3.82)}   & 63,847,601 (8.58$\times$) & \multicolumn{1}{c}{56.71\% (-20.12)} \\ \midrule
& \texttt{normal} & 653,476    & \multicolumn{1}{c|}{63.33\%} & 6,073,627    & \multicolumn{1}{c|}{70.05\%} & 5,369,616 & \multicolumn{1}{c}{60.98\%} \\
\multirow{-2}{*}{\textbf{Qwen3-Thinking}}   & \texttt{mixed}  & 2,501,360 (3.83$\times$)  & \multicolumn{1}{c|}{26.67\% (-36.66)}    & 28,604,299 (4.71$\times$)   & \multicolumn{1}{c|}{64.97\% (-5.08)}   & 14,327,613 (2.67$\times$) & \multicolumn{1}{c}{52.44\% (-8.54)}  \\ \midrule
& \texttt{normal} & 5,051,562  & \multicolumn{1}{c|}{20.00\%} & 45,622,515   & \multicolumn{1}{c|}{47.21\%} & 12,419,795 & \multicolumn{1}{c}{30.49\%} \\
\multirow{-2}{*}{\textbf{Qwen3-Coder}}      & \texttt{mixed}  & 15,689,070 (3.11$\times$) & \multicolumn{1}{c|}{23.33\% (+3.33)}     & 114,800,037 (2.52$\times$)  & \multicolumn{1}{c|}{48.22\% (+1.01)}   & 87,259,072 (7.03$\times$) & \multicolumn{1}{c}{3.05\% (-27.44)}  \\ \midrule
& \texttt{normal} & 549,667    & \multicolumn{1}{c|}{60.00\%} & 2,752,992    & \multicolumn{1}{c|}{70.05\%} & 2,361,707 & \multicolumn{1}{c}{32.32\%} \\
\multirow{-2}{*}{\textbf{DeepSeek-V3.2}}    & \texttt{mixed}  & 4,948,229 (9.00$\times$)  & \multicolumn{1}{c|}{23.33\% (-36.67)}    & 11,141,346 (4.05$\times$)   & \multicolumn{1}{c|}{26.90\% (-43.15)}  & 10,751,180 (4.55$\times$) & \multicolumn{1}{c}{6.71\% (-25.61)}  \\ \midrule
& \texttt{normal} & 6,358,821  & \multicolumn{1}{c|}{40.00\%} & 10,272,416   & \multicolumn{1}{c|}{58.88\%} & 6,508,864 & \multicolumn{1}{c}{40.85\%} \\
\multirow{-2}{*}{\textbf{GLM-4.7}}  & \texttt{mixed}  & 17,464,231 (2.75$\times$) & \multicolumn{1}{c|}{60.00\% (+20.00)}    & 110,588,180 (10.77$\times$) & \multicolumn{1}{c|}{53.81\% (-5.07)}   & 94,935,247 (14.59$\times$) & \multicolumn{1}{c}{9.15\% (-31.70)}  \\
\bottomrule
\end{tabular}%
}
\caption{
\textbf{Accuracy and token consumption in \textit{ReAct}-based agent settings.}
Results are reported under \texttt{normal} and \texttt{mixed} tool registry compositions for AIME2025, GPQA Diamond, and HumanEval.
Parentheses report the $\times$-factor relative to \texttt{normal}. 
}
\label{tab:react-wo-nowait}
\end{table*}

%% file: table_latex/revision_qwen_stats.tex
\begin{table*}[t]
\centering
\scriptsize
\setlength{\tabcolsep}{3.5pt}
\renewcommand{\arraystretch}{0.9}
\resizebox{\textwidth}{!}{%
\begin{tabular}{llrrrrrrrrr}
\toprule
\textbf{Model} & \textbf{Registry} & \textbf{$n$} &
\textbf{Total Mean} & \textbf{Total Min} & \textbf{Total Max} &
\textbf{Input Mean} & \textbf{Output Mean} &
\textbf{Active Mean} & \textbf{Active Max} & \textbf{Acc.} \\
\midrule
\multirow{2}{*}{\textbf{Kimi-K2-Thinking}}
& \texttt{normal} & 5 & 4,051,860 & 4,009,470 & 4,132,093 & 3,979,866 & 71,994 & 2,575.80 & 3,425.11 & 75.98 \\
& \texttt{mixed} & 5 & 60,733,376 & 57,566,674 & 64,217,917 & 60,158,105 & 575,271 & 22,420.17 & 32,368.59 & 76.46 \\
\midrule
\multirow{2}{*}{\textbf{MiniMax-M2}}
& \texttt{normal} & 5 & 4,918,607 & 4,909,299 & 4,930,458 & 4,761,566 & 157,042 & 6,165.84 & 6,171.32 & 86.38 \\
& \texttt{mixed} & 5 & 99,651,664 & 91,711,604 & 112,484,616 & 98,918,351 & 733,313 & 31,048.17 & 34,881.23 & 66.26 \\
\midrule
\multirow{2}{*}{\textbf{Qwen3-Thinking}}
& \texttt{normal} & 5 & 7,619,068 & 6,948,487 & 8,252,654 & 7,483,692 & 135,377 & 4,879.75 & 6,769.51 & 76.83 \\
& \texttt{mixed} & 5 & 59,427,673 & 58,162,299 & 61,668,627 & 58,864,769 & 562,904 & 17,809.37 & 21,956.87 & 54.39 \\
\midrule
\multirow{2}{*}{\textbf{Qwen3-Coder}}
& \texttt{normal} & 5 & 9,819,312 & 7,076,806 & 11,305,864 & 9,695,476 & 123,836 & 10,151.47 & 16,108.71 & 83.33 \\
& \texttt{mixed} & 5 & 42,150,846 & 17,210,280 & 61,605,432 & 41,683,419 & 467,427 & 6,334.57 & 9,242.07 & 13.21 \\
\midrule
\multirow{2}{*}{\textbf{DeepSeek-V3.2}}
& \texttt{normal} & 5 & 12,032,964 & 11,376,239 & 12,440,445 & 11,827,419 & 205,545 & 15,820.91 & 26,175.76 & 86.99 \\
& \texttt{mixed} & 5 & 105,245,873 & 25,905,567 & 151,910,916 & 104,184,303 & 1,061,570 & 17,211.48 & 29,402.16 & 24.59 \\
\midrule
\multirow{2}{*}{\textbf{GLM-4.7}}
& \texttt{normal} & 5 & 7,469,745 & 7,350,801 & 7,668,691 & 7,347,776 & 121,969 & 11,102.60 & 13,942.49 & 95.12 \\
& \texttt{mixed} & 5 & 50,776,468 & 38,206,084 & 75,702,975 & 50,217,756 & 558,712 & 15,080.85 & 31,502.00 & 23.78 \\
\bottomrule
\end{tabular}%
}
\caption{
\textbf{\textit{Qwen-Code} accounting statistics on full HumanEval.}
Statistics are run-level aggregates under each registry configuration. Total, input, and output columns report token counts; active-time columns report aggregate active execution time in seconds. The wall-clock cap is applied per problem instance, not to the full benchmark run.
}
\label{tab:qwen-code-accounting-stats}
\end{table*}

%% file: table_latex/policy.tex
\begin{table*}[hbpt]
\centering
\scriptsize
\resizebox{\textwidth}{!}{%
\begin{tabular}{l|l|l|c}
\toprule
\textbf{Model} & \textbf{Provider} & \textbf{Key Restrictions for Research} & \textbf{Source (Access Date)} \\ 
\midrule
GPT-4o & OpenAI & Prohibition on using output to develop competing models & \href{https://openai.com/policies/row-terms-of-use/}{Link} (2026-05-25) \\
Claude 3.5 & Anthropic & Prohibition on automated data scraping and extraction & \href{https://www.anthropic.com/legal/consumer-terms}{Link} (2026-05-25) \\
Gemini & Google & Regional availability and specific API usage quotas & \href{https://ai.google.dev/gemini-api/terms}{Link} (2026-05-25) \\
\bottomrule
\end{tabular}
}
\caption{
\textbf{Summary of proprietary model terms and conditions. }
Each link provides the official documentation used to determine experimental constraints.
}
\label{tab:model-terms-appendix}
\vspace{-1.0em}
\end{table*}

%% file: custom.bib
@article{yang2026mcpsecbenchsystematicsecuritybenchmark,
  title={{MCPSecBench}: A systematic security benchmark and playground for testing {Model Context Protocols}},
  author={Yang, Yixuan and Gao, Cuifeng and Wu, Daoyuan and Chen, Yufan and Li, Yingjiu and Wang, Shuai},
  journal={arXiv preprint arXiv:2508.13220},
  year={2025}
}

@inproceedings{rein2024gpqa,
    title={{GPQA}: A Graduate-Level Google-Proof Q\&A Benchmark},
    author={David Rein and Betty Li Hou and Asa Cooper Stickland and Jackson Petty and Richard Yuanzhe Pang and Julien Dirani and Julian Michael and Samuel R. Bowman},
    booktitle={First Conference on Language Modeling},
    year={2024},
    url={https://openreview.net/forum?id=Ti67584b98}
}

@inproceedings{shi2025retrieval,
  title={Retrieval models aren’t tool-savvy: Benchmarking tool retrieval for large language models},
  author={Shi, Zhengliang and Wang, Yuhan and Yan, Lingyong and Ren, Pengjie and Wang, Shuaiqiang and Yin, Dawei and Ren, Zhaochun},
  booktitle={Findings of the Association for Computational Linguistics: ACL 2025},
  pages={24497--24524},
  year={2025}
}

@misc{zhou2026stealthy,
      title={Beyond Max Tokens: Stealthy Resource Amplification via Tool Calling Chains in LLM Agents}, 
      author={Kaiyu Zhou and Yongsen Zheng and Yicheng He and Meng Xue and Xueluan Gong and Yuji Wang and Kwok-Yan Lam},
      year={2026},
      eprint={2601.10955},
      archivePrefix={arXiv},
      primaryClass={cs.CR},
      url={https://arxiv.org/abs/2601.10955}, 
}

@inproceedings{wang2025mcptox,
  title={{MCPTox}: A benchmark for tool poisoning on real-world {MCP} servers},
  author={Wang, Zhiqiang and Gao, Yichao and Wang, Yanting and Liu, Suyuan and Sun, Haifeng and Cheng, Haoran and Shi, Guanquan and Du, Haohua and Li, Xiangyang},
  booktitle={Proceedings of the AAAI Conference on Artificial Intelligence},
  volume={40},
  number={42},
  pages={35811--35819},
  year={2026}
}

@article{schick2023toolformerlanguagemodelsteach,
  title={{Toolformer}: Language models can teach themselves to use tools},
  author={Schick, Timo and Dwivedi-Yu, Jane and Dess{\`\i}, Roberto and Raileanu, Roberta and Lomeli, Maria and Hambro, Eric and Zettlemoyer, Luke and Cancedda, Nicola and Scialom, Thomas},
  journal={Advances in neural information processing systems},
  volume={36},
  pages={68539--68551},
  year={2023}
}

@inproceedings{qin2023toolllmfacilitatinglargelanguage,
  title={{ToolLLM}: Facilitating large language models to master 16000+ real-world {APIs}},
  author={Qin, Yujia and Liang, Shihao and Ye, Yining and Zhu, Kunlun and Yan, Lan and Lu, Yaxi and Lin, Yankai and Cong, Xin and Tang, Xiangru and Qian, Bill and others},
  booktitle={International Conference on Learning Representations},
  volume={2024},
  pages={9695--9717},
  year={2024}
}

@inproceedings{
    chen2025do,
    title={Do {NOT} Think That Much for 2+3=? On the Overthinking of Long Reasoning Models},
    author={Xingyu Chen and Jiahao Xu and Tian Liang and Zhiwei He and Jianhui Pang and Dian Yu and Linfeng Song and Qiuzhi Liu and Mengfei Zhou and Zhuosheng Zhang and Rui Wang and Zhaopeng Tu and Haitao Mi and Dong Yu},
    booktitle={Forty-second International Conference on Machine Learning},
    year={2025},
    url={https://openreview.net/forum?id=MSbU3L7V00}
}

@article{sui2025stopoverthinkingsurveyefficient,
  title={Stop overthinking: A survey on efficient reasoning for large language models},
  author={Sui, Yang and Chuang, Yu-Neng and Wang, Guanchu and Zhang, Jiamu and Zhang, Tianyi and Yuan, Jiayi and Liu, Hongyi and Wen, Andrew and Zhong, Shaochen and Zou, Na and others},
  journal={arXiv preprint arXiv:2503.16419},
  year={2025}
}

@misc{su2025underthinkingoverthinkingempiricalstudy,
      title={Between Underthinking and Overthinking: An Empirical Study of Reasoning Length and correctness in LLMs}, 
      author={Jinyan Su and Jennifer Healey and Preslav Nakov and Claire Cardie},
      year={2025},
      eprint={2505.00127},
      archivePrefix={arXiv},
      primaryClass={cs.CL},
      url={https://arxiv.org/abs/2505.00127}, 
}

@misc{cuadron2025dangeroverthinkingexaminingreasoningaction,
      title={The Danger of Overthinking: Examining the Reasoning-Action Dilemma in Agentic Tasks}, 
      author={Alejandro Cuadron and Dacheng Li and Wenjie Ma and Xingyao Wang and Yichuan Wang and Siyuan Zhuang and Shu Liu and Luis Gaspar Schroeder and Tian Xia and Huanzhi Mao and Nicholas Thumiger and Aditya Desai and Ion Stoica and Ana Klimovic and Graham Neubig and Joseph E. Gonzalez},
      year={2025},
      eprint={2502.08235},
      archivePrefix={arXiv},
      primaryClass={cs.AI},
      url={https://arxiv.org/abs/2502.08235}, 
}

@misc{hassid2025dontoverthinkitpreferring,
      title={Don't Overthink it. Preferring Shorter Thinking Chains for Improved LLM Reasoning}, 
      author={Michael Hassid and Gabriel Synnaeve and Yossi Adi and Roy Schwartz},
      year={2025},
      eprint={2505.17813},
      archivePrefix={arXiv},
      primaryClass={cs.CL},
      url={https://arxiv.org/abs/2505.17813}, 
}

@inproceedings{wang-etal-2025-wait,
    title = "Wait, We Don{'}t Need to ``Wait''! Removing Thinking Tokens Improves Reasoning Efficiency",
    author = "Wang, Chenlong  and
      Feng, Yuanning  and
      Chen, Dongping  and
      Chu, Zhaoyang  and
      Krishna, Ranjay  and
      Zhou, Tianyi",
    editor = "Christodoulopoulos, Christos  and
      Chakraborty, Tanmoy  and
      Rose, Carolyn  and
      Peng, Violet",
    booktitle = "Findings of the Association for Computational Linguistics: EMNLP 2025",
    month = nov,
    year = "2025",
    address = "Suzhou, China",
    publisher = "Association for Computational Linguistics",
    url = "https://aclanthology.org/2025.findings-emnlp.394/",
    doi = "10.18653/v1/2025.findings-emnlp.394",
    pages = "7459--7482",
    ISBN = "979-8-89176-335-7"
}

@inproceedings{wei2022cot,
    author = {Wei, Jason and Wang, Xuezhi and Schuurmans, Dale and Bosma, Maarten and Ichter, Brian and Xia, Fei and Chi, Ed H. and Le, Quoc V. and Zhou, Denny},
    title = {Chain-of-thought prompting elicits reasoning in large language models},
    year = {2022},
    isbn = {9781713871088},
    publisher = {Curran Associates Inc.},
    address = {Red Hook, NY, USA},
    booktitle = {Proceedings of the 36th International Conference on Neural Information Processing Systems},
    articleno = {1800},
    numpages = {14},
    location = {New Orleans, LA, USA},
    series = {NIPS '22}
}

@inproceedings{kojima2022large,
    author = {Kojima, Takeshi and Gu, Shixiang Shane and Reid, Machel and Matsuo, Yutaka and Iwasawa, Yusuke},
    title = {Large language models are zero-shot reasoners},
    year = {2022},
    isbn = {9781713871088},
    publisher = {Curran Associates Inc.},
    address = {Red Hook, NY, USA},
    booktitle = {Proceedings of the 36th International Conference on Neural Information Processing Systems},
    articleno = {1613},
    numpages = {15},
    location = {New Orleans, LA, USA},
    series = {NIPS '22}
}

@inproceedings{
    wang2023selfconsistency,
    title={Self-Consistency Improves Chain of Thought Reasoning in Language Models},
    author={Xuezhi Wang and Jason Wei and Dale Schuurmans and Quoc V Le and Ed H. Chi and Sharan Narang and Aakanksha Chowdhery and Denny Zhou},
    booktitle={The Eleventh International Conference on Learning Representations },
    year={2023},
    url={https://openreview.net/forum?id=1PL1NIMMrw}
}

@inproceedings{yao2023treeofthought,
    author = {Yao, Shunyu and Yu, Dian and Zhao, Jeffrey and Shafran, Izhak and Griffiths, Thomas L. and Cao, Yuan and Narasimhan, Karthik},
    title = {Tree of thoughts: deliberate problem solving with large language models},
    year = {2023},
    publisher = {Curran Associates Inc.},
    address = {Red Hook, NY, USA},
    booktitle = {Proceedings of the 37th International Conference on Neural Information Processing Systems},
    articleno = {517},
    numpages = {14},
    location = {New Orleans, LA, USA},
    series = {NIPS '23}
}

@inproceedings{
    lightman2024lets,
    title={Let's Verify Step by Step},
    author={Hunter Lightman and Vineet Kosaraju and Yuri Burda and Harrison Edwards and Bowen Baker and Teddy Lee and Jan Leike and John Schulman and Ilya Sutskever and Karl Cobbe},
    booktitle={The Twelfth International Conference on Learning Representations},
    year={2024},
    url={https://openreview.net/forum?id=v8L0pN6EOi}
}

@inproceedings{10.1145/3605764.3623985,
    author = {Greshake, Kai and Abdelnabi, Sahar and Mishra, Shailesh and Endres, Christoph and Holz, Thorsten and Fritz, Mario},
    title = {Not What You've Signed Up For: Compromising Real-World LLM-Integrated Applications with Indirect Prompt Injection},
    year = {2023},
    isbn = {9798400702600},
    publisher = {Association for Computing Machinery},
    address = {New York, NY, USA},
    url = {https://doi.org/10.1145/3605764.3623985},
    doi = {10.1145/3605764.3623985},
    booktitle = {Proceedings of the 16th ACM Workshop on Artificial Intelligence and Security},
    pages = {79–90},
    numpages = {12},
    location = {Copenhagen, Denmark},
    series = {AISec '23}
}

@inproceedings{zou2025poisonedrag,
  title={$\{$PoisonedRAG$\}$: Knowledge corruption attacks to $\{$Retrieval-Augmented$\}$ generation of large language models},
  author={Zou, Wei and Geng, Runpeng and Wang, Binghui and Jia, Jinyuan},
  booktitle={34th USENIX Security Symposium (USENIX Security 25)},
  pages={3827--3844},
  year={2025}
}

@misc{aime2025,
  author = {{Mathematical Association of America}},
  title  = {American Invitational Mathematics Examination -- AIME},
  year   = {2025},
  month  = feb,
  howpublished = {American Invitational Mathematics Examination -- AIME 2025},
  url    = {https://maa.org/maa-invitational-competitions/}
}

@misc{chen2021codex,
  title={Evaluating Large Language Models Trained on Code},
  author={Mark Chen and Jerry Tworek and Heewoo Jun and Qiming Yuan and Henrique Ponde de Oliveira Pinto and Jared Kaplan and Harri Edwards and Yuri Burda and Nicholas Joseph and Greg Brockman and Alex Ray and Raul Puri and Gretchen Krueger and Michael Petrov and Heidy Khlaaf and Girish Sastry and Pamela Mishkin and Brooke Chan and Scott Gray and Nick Ryder and Mikhail Pavlov and Alethea Power and Lukasz Kaiser and Mohammad Bavarian and Clemens Winter and Philippe Tillet and Felipe Petroski Such and Dave Cummings and Matthias Plappert and Fotios Chantzis and Elizabeth Barnes and Ariel Herbert-Voss and William Hebgen Guss and Alex Nichol and Alex Paino and Nikolas Tezak and Jie Tang and Igor Babuschkin and Suchir Balaji and Shantanu Jain and William Saunders and Christopher Hesse and Andrew N. Carr and Jan Leike and Josh Achiam and Vedant Misra and Evan Morikawa and Alec Radford and Matthew Knight and Miles Brundage and Mira Murati and Katie Mayer and Peter Welinder and Bob McGrew and Dario Amodei and Sam McCandlish and Ilya Sutskever and Wojciech Zaremba},
  year={2021},
  eprint={2107.03374},
  archivePrefix={arXiv},
  primaryClass={cs.LG},
  url={https://arxiv.org/abs/2107.03374}
}

@misc{5team2025glm45agenticreasoningcoding,
      title={GLM-4.5: Agentic, Reasoning, and Coding (ARC) Foundation Models}, 
      author={{GLM Team}},
      year={2025},
      eprint={2508.06471},
      archivePrefix={arXiv},
      primaryClass={cs.CL},
      url={https://arxiv.org/abs/2508.06471}, 
}

@article{deepseekai2025deepseekv32,
  title={{DeepSeek-V3.2}: Pushing the frontier of open large language models},
  author={Liu, Aixin and Mei, Aoxue and Lin, Bangcai and Xue, Bing and Wang, Bingxuan and Xu, Bingzheng and Wu, Bochao and Zhang, Bowei and Lin, Chaofan and Dong, Chen and others},
  journal={arXiv preprint arXiv:2512.02556},
  year={2025}
}

@misc{qwen3technicalreport,
      title={Qwen3 Technical Report}, 
      author={{Qwen Team}},
      year={2025},
      eprint={2505.09388},
      archivePrefix={arXiv},
      primaryClass={cs.CL},
      url={https://arxiv.org/abs/2505.09388}, 
}

@misc{minimax2025m2,
  author = {{MiniMax AI}},
  title  = {{MiniMax-M2}},
  year   = {2025},
  url    = {https://huggingface.co/MiniMaxAI/MiniMax-M2}
}

@misc{moonshot2025k2thinking,
  author = {{Moonshot AI}},
  title  = {{Kimi K2-Thinking}},
  year   = {2025},
  url    = {https://moonshotai.github.io/Kimi-K2/thinking.html}
}

@inproceedings{
dong2025an,
title={An Engorgio Prompt Makes Large Language Model Babble on},
author={Jianshuo Dong and Ziyuan Zhang and Qingjie Zhang and Tianwei Zhang and Hao Wang and Hewu Li and Qi Li and Chao Zhang and Ke Xu and Han Qiu},
booktitle={The Thirteenth International Conference on Learning Representations},
year={2025},
url={https://openreview.net/forum?id=m4eXBo0VNc}
}

@misc{gao2024denialofservicepoisoningattackslarge,
      title={Denial-of-Service Poisoning Attacks against Large Language Models}, 
      author={Kuofeng Gao and Tianyu Pang and Chao Du and Yong Yang and Shu-Tao Xia and Min Lin},
      year={2024},
      eprint={2410.10760},
      archivePrefix={arXiv},
      primaryClass={cs.CR},
      url={https://arxiv.org/abs/2410.10760}, 
}

@inproceedings{zhang-etal-2025-crabs,
    title = "Crabs: Consuming Resource via Auto-generation for {LLM}-{D}o{S} Attack under Black-box Settings",
    author = "Zhang, Yuanhe  and
      Zhou, Zhenhong  and
      Zhang, Wei  and
      Wang, Xinyue  and
      Jia, Xiaojun  and
      Liu, Yang  and
      Su, Sen",
    editor = "Che, Wanxiang  and
      Nabende, Joyce  and
      Shutova, Ekaterina  and
      Pilehvar, Mohammad Taher",
    booktitle = "Findings of the Association for Computational Linguistics: ACL 2025",
    month = jul,
    year = "2025",
    address = "Vienna, Austria",
    publisher = "Association for Computational Linguistics",
    url = "https://aclanthology.org/2025.findings-acl.580/",
    doi = "10.18653/v1/2025.findings-acl.580",
    pages = "11128--11150",
    ISBN = "979-8-89176-256-5"
}

@misc{wan2026mitigatingoverthinkinglargereasoning,
      title={Mitigating Overthinking in Large Reasoning Models via Difficulty-aware Reinforcement Learning}, 
      author={Qian Wan and Ziao Xu and Luona Wei and Xiaoxuan Shen and Jianwen Sun},
      year={2026},
      eprint={2601.21418},
      archivePrefix={arXiv},
      primaryClass={cs.LG},
      url={https://arxiv.org/abs/2601.21418}, 
}

@inproceedings{yao2023react,
  title = {{ReAct}: Synergizing Reasoning and Acting in Language Models},
  author = {Yao, Shunyu and Zhao, Jeffrey and Yu, Dian and Du, Nan and Shafran, Izhak and Narasimhan, Karthik and Cao, Yuan},
  booktitle = {International Conference on Learning Representations (ICLR) },
  year = {2023},
  html = {https://arxiv.org/abs/2210.03629},
}

@inproceedings{lu-etal-2025-toolsandbox,
    title = "{T}ool{S}andbox: A Stateful, Conversational, Interactive Evaluation Benchmark for {LLM} Tool Use Capabilities",
    author = "Lu, Jiarui  and
      Holleis, Thomas  and
      Zhang, Yizhe  and
      Aumayer, Bernhard  and
      Nan, Feng  and
      Bai, Haoping  and
      Ma, Shuang  and
      Ma, Shen  and
      Li, Mengyu  and
      Yin, Guoli  and
      Wang, Zirui  and
      Pang, Ruoming",
    editor = "Chiruzzo, Luis  and
      Ritter, Alan  and
      Wang, Lu",
    booktitle = "Findings of the Association for Computational Linguistics: NAACL 2025",
    month = apr,
    year = "2025",
    address = "Albuquerque, New Mexico",
    publisher = "Association for Computational Linguistics",
    url = "https://aclanthology.org/2025.findings-naacl.65/",
    doi = "10.18653/v1/2025.findings-naacl.65",
    pages = "1160--1183",
    ISBN = "979-8-89176-195-7"
}

@article{FERRAG2025,
  title={From prompt injections to protocol exploits: Threats in {LLM}-powered {AI} agents workflows},
  author={Ferrag, Mohamed Amine and Tihanyi, Norbert and Hamouda, Djallel and Maglaras, Leandros and Lakas, Abderrahmane and Debbah, Merouane},
  journal={ICT Express},
  year={2025},
  publisher={Elsevier}
}

@inproceedings{zhang-etal-2025-allies,
    title = "From Allies to Adversaries: Manipulating {LLM} Tool-Calling through Adversarial Injection",
    author = "Zhang, Rupeng  and
      Wang, Haowei  and
      Wang, Junjie  and
      Li, Mingyang  and
      Huang, Yuekai  and
      Wang, Dandan  and
      Wang, Qing",
    editor = "Chiruzzo, Luis  and
      Ritter, Alan  and
      Wang, Lu",
    booktitle = "Proceedings of the 2025 Conference of the Nations of the Americas Chapter of the Association for Computational Linguistics: Human Language Technologies (Volume 1: Long Papers)",
    month = apr,
    year = "2025",
    address = "Albuquerque, New Mexico",
    publisher = "Association for Computational Linguistics",
    url = "https://aclanthology.org/2025.naacl-long.101/",
    doi = "10.18653/v1/2025.naacl-long.101",
    pages = "2009--2028",
    ISBN = "979-8-89176-189-6"
}

@misc{george2025unsupervisedcycledetectionagentic,
      title={Unsupervised Cycle Detection in Agentic Applications}, 
      author={Felix George and Harshit Kumar and Divya Pathak and Kaustabha Ray and Mudit Verma and Pratibha Moogi},
      year={2025},
      eprint={2511.10650},
      archivePrefix={arXiv},
      primaryClass={cs.CL},
      url={https://arxiv.org/abs/2511.10650}, 
}

@misc{zeng2026tooloveruseillusiondoesllm,
      title={The Tool-Overuse Illusion: Why Does LLM Prefer External Tools over Internal Knowledge?}, 
      author={Yirong Zeng and Shen You and Yufei Liu and Qunyao Du and Xiao Ding and Yutai Hou and Yuxian Wang and Wu Ning and Haonan Song and Dandan Tu and Bibo Cai and Ting Liu},
      year={2026},
      eprint={2604.19749},
      archivePrefix={arXiv},
      primaryClass={cs.AI},
      url={https://arxiv.org/abs/2604.19749}, 
}

@misc{huang2026modelcontextprotocolthreat,
      title={Model Context Protocol Threat Modeling and Analyzing Vulnerabilities to Prompt Injection with Tool Poisoning}, 
      author={Charoes Huang and Xin Huang and Ngoc Phu Tran and Amin Milani Fard},
      year={2026},
      eprint={2603.22489},
      archivePrefix={arXiv},
      primaryClass={cs.CR},
      url={https://arxiv.org/abs/2603.22489}, 
}

@misc{hou2026smcpsecuremodelcontext,
      title={SMCP: Secure Model Context Protocol}, 
      author={Xinyi Hou and Shenao Wang and Yifan Zhang and Ziluo Xue and Yanjie Zhao and Cai Fu and Haoyu Wang},
      year={2026},
      eprint={2602.01129},
      archivePrefix={arXiv},
      primaryClass={cs.CR},
      url={https://arxiv.org/abs/2602.01129}, 
}

@misc{xu2026evolutiontoolusellm,
      title={The Evolution of Tool Use in LLM Agents: From Single-Tool Call to Multi-Tool Orchestration}, 
      author={Haoyuan Xu and Chang Li and Xinyan Ma and Xianhao Ou and Zihan Zhang and Tao He and Xiangyu Liu and Zixiang Wang and Jiafeng Liang and Zheng Chu and Runxuan Liu and Rongchuan Mu and Dandan Tu and Ming Liu and Bing Qin},
      year={2026},
      eprint={2603.22862},
      archivePrefix={arXiv},
      primaryClass={cs.SE},
      url={https://arxiv.org/abs/2603.22862}, 
}
